\documentclass[letterpaper]{article} 
\usepackage{aaai2026}  
\usepackage{times}  
\usepackage{helvet}  
\usepackage{courier}  
\usepackage[hyphens]{url}  
\usepackage{graphicx} 
\usepackage{natbib}  
\usepackage{caption} 
\usepackage{algorithm}
\usepackage{algpseudocode}
\usepackage{amsmath}
\usepackage{amsfonts}

\usepackage{amsthm}
\usepackage{xcolor}
\usepackage{color}
\usepackage{soul}

\usepackage{booktabs}  
\usepackage{multirow}

\newtheorem{defi}{Definition}
\newtheorem{remark}{Remark}
\newtheorem{theorem}{Theorem}
\newtheorem{lemma}{Lemma}
\newtheorem{assum}{Assumption}

\usepackage{newfloat}
\usepackage{listings}
\DeclareCaptionStyle{ruled}{labelfont=normalfont,labelsep=colon,strut=off} 
\floatstyle{ruled}
\newfloat{listing}{tb}{lst}{}
\floatname{listing}{Listing}
\DeclareMathSizes{10}{8}{6}{5}

\nocopyright

\title{Everyone Contributes! Incentivizing Strategic Cooperation in \\ Multi-LLM Systems via Sequential Public Goods Games}
\author{
    Yunhao Liang\textsuperscript{\rm 1},
    Yuan Qu\textsuperscript{\rm 2}\thanks{Correspondence to: Yuan Qu (yuanqu@hku.hk), Jingyuan Yang (jyang53@gmu.edu)}, Jingyuan Yang\textsuperscript{\rm 3\textasteriskcentered{}}, Shaochong Lin\textsuperscript{\rm 2}, Zuo-Jun Max Shen\textsuperscript{\rm 2 4 5}
}
\affiliations{
    \textsuperscript{\rm 1}The University of Hong Kong, Shenzhen Institute of Research and Innovation, Shenzhen, Guangdong, China\\
    \textsuperscript{\rm 2}Faculty of Engineering, The University of Hong Kong, Hong Kong, China\\
    \textsuperscript{\rm 3}Costello College of Business, George Mason University, Fairfax, VA 22030, USA\\ 
    \textsuperscript{\rm 4}Faculty of Business and Economics, The University of Hong Kong, Hong Kong, China\\
    \textsuperscript{\rm 5}College of Engineering, University of California, Berkeley, CA 94720, USA\\

}

\begin{document}

\maketitle

\begin{abstract}

Coordinating multiple large language models (LLMs) to solve complex tasks collaboratively poses a fundamental trade‑off between the computation costs and collective performance compared with
individual model. We introduce a novel, game‑theoretically grounded reinforcement learning (RL) framework, the Multi-Agent Cooperation Sequential Public Goods Game (MAC-SPGG), to systematically incentivize cooperation in multi‑LLM ensembles. In MAC-SPGG, LLM agents move in sequence, observing predecessors’ outputs and updating beliefs to condition their own contributions. By redesigning the public‑goods reward, effortful contributions become the unique Subgame Perfect Nash Equilibrium (SPNE), which eliminates free‑riding under traditional SPGG or PGG. Its sequential protocol replaces costly round‑based information exchanges with a streamlined decision flow, cutting communication overhead while retaining strategic depth. We prove the existence and uniqueness of the SPNE under realistic parameters, and empirically show that MAC-SPGG-trained ensembles outperform single‑agent baselines, chain‑of‑thought prompting, and other cooperative methods, even achieving comparable performance to large-scale models across reasoning, math, code generation, and NLP tasks. Our results highlight the power of structured, incentive-aligned MAC-SPGG cooperation for scalable and robust multi-agent language generation.

\end{abstract}

\section{Introduction}

Recent advancements in large language models (LLMs) have demonstrated impressive capabilities across various reasoning and decision-making tasks, especially within multi-agent scenarios. Emerging research~\cite{llmsSurvey} explores diverse interaction paradigms among multiple LLMs, from competitive debating and strategic reasoning~\cite{cheng2024selfplaying, du2024multiagentDebate, LEGO, liang2024encouragingDebate, Yi2025FromDT} to cooperative decision-making and collaborative problem-solving~\cite{li2024moreNeed, li2023camel, hong2024metagpt, quan2025invagent, yao2025comal, chen2024reconcile}. Multi-LLM ensembles are promising because they combine complementary reasoning strategies, diversify knowledge sources, and improve robustness and accuracy over single-model systems.

However, achieving these benefits crucially depends on effectively coordinating the ensemble, especially from the information-sharing perspective. Existing frameworks predominantly rely on two communication strategies: simultaneous and sequential. In the simultaneous setting, LLMs act independently and concurrently, requiring a central coordinator to aggregate outputs. This single-point bottleneck raises communication cost and limits dynamic, information-driven interaction within the ensemble~\cite{Hammond2025MultiAgentRisk, ECON}.
Conversely, sequential communication enables information sharing among agents, allowing each model to condition its action on preceding outputs. 
However, without careful strategic design, unrestricted sequential information exchange accumulated among all agents can lead to significant communication overhead and computational complexity~\cite{Whymultifail, Hiddenprofile, InfoAsymmetry}.

Hence, a critical challenge arises: \textit{how can we achieve high-performance multi-LLM ensembles 
while reducing communication and computational overhead?} Inspired by game theory, where all the players contribute rationally with only a common knowledge of the game rules, we adopt the idea of Public Goods Game (PGG) for the multi-LLM ensemble learning. PGG is a canonical paradigm extensively examined in economics and behavioral sciences~\cite{fehr2002altruistic, Anwar2023PositionSPGG}, which characterizes scenarios where individuals contribute to a collective good, balancing private costs against shared public benefits. 
Prominent real-world examples include crowdfunding platforms~\cite{Belleflamme2013CrowdfundingTT}, open-source collaborations~\cite{Tirole2002SomeOpenSource, Forte2005WhyDPWikipedia}, and public infrastructure funded by taxation~\cite{Connolly1999EconomicsOT}.

\begin{figure}[t!]
    \centering
    \includegraphics[width=0.8\linewidth]{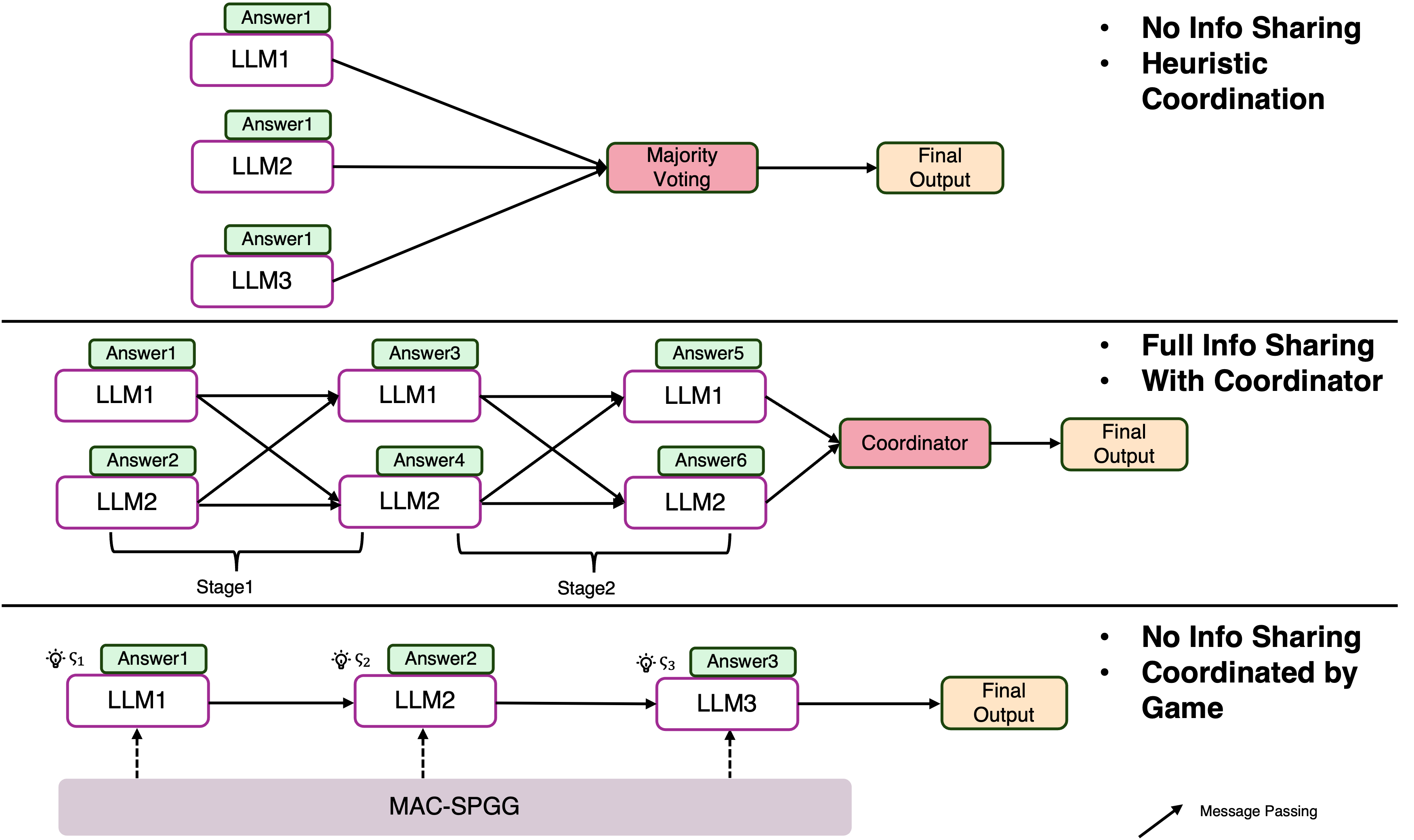}
    \caption{Comparison of coordination mechanisms across LLM-based multi-agent systems.}
    \label{fig:flow_compare}
\vspace{-4ex}
\end{figure}

Building upon this paradigm, we propose the two-phase game-theoretical reinforcement learning (RL) framework, \textit{Multi-Agent Cooperation Sequential Public Goods Game (MAC-SPGG)}, as a theoretical foundation to coordinate multi-LLM ensembles systematically. 
While SPGG is established in the game-theory literature~\cite{Anwar2023PositionSPGG,Anwar2023PositionUI,gachter2010sequential}, its implications for LLM ensembles remain underexplored.
Our MAC-SPGG explicitly models sequential decision-making, where agents observe predecessors' contributions before acting—a scenario naturally aligning with multi-LLM frameworks such as cascading prompting~\cite{zhang2024prefer} and iterative refinement~\cite{chen2024magicore}. 
Different from the existing coordinator-based multi-LLM ensembles,
MAC-SPGG enables each model to evaluate prior contributions sequentially. Its carefully designed reward structure motivates each model to participate positively, promoting stable cooperative equilibria and ultimately enhancing ensemble performance; see Figure~\ref{fig:flow_compare}.
By incentivizing the sequential coordination process, MAC-SPGG eliminates the central coordinator, substantially reduces associated costs, and strengthens collaboration among agents.
In this paper, we formally develop and validate the effectiveness of our SPGG-based multi-LLM coordination mechanism.

In our framework, we prove that a unique \textit{Subgame Perfect Nash Equilibrium (SPNE)} can be found under reasonable conditions in the inference phase, the SPGG part.
By adjusting the incentives in traditional PGG, the equilibrium shifts from free-riding to positively cooperative participation. Our theoretically guaranteed equilibrium behaviors are largely absent from existing debate-, voting-, or heuristic-based coordination methods~\cite{du2024multiagentDebate,li2024moreNeed, Chen2023AgentVerseFM, Chen2023ReConcileRC}. 
Our framework significantly reduces communication overhead compared to iterative information exchanges, while preserving strategic depth. 


In the optimization phase, the learning part, our training process demonstrates its power empirically. In experiments, MAC-SPGG robustly directs multi-LLM ensembles toward cooperative equilibria, consistently outperforming single-agent baselines, Chain-of-Thought (CoT) prompting~\cite{Wei2022COT}, and other cooperative frameworks across four diverse tasks, including code generation (HumanEval), factual knowledge (MMLU), mathematical reasoning (GSM8K), and natural language understanding (SummEval). We systematically assess two Bayesian belief update strategies, \textit{Partial Observation (PO) and Full Observation (FO)}, reflecting varying levels of inter-agent information transparency. 

Our key contributions are summarized as follows:
\begin{itemize}
    \item We propose a theoretically grounded MAC-SPGG framework for structured multi-LLM cooperation. The existence and uniqueness of the SPNE provide theoretical foundations for equilibrium-driven cooperation.
    \item We empirically test MAC-SPGG across varied tasks and ablation tests, which consistently outperforms other single-agent and cooperative benchmarks. We find that optimal information sharing is context-dependent, and minimal transparency may yield superior outcomes.
\end{itemize}

\section{Related Work}\label{sec:literature}

Our work synthesizes insights from multi-agent collaboration and mechanism design in LLM systems.

\noindent\textbf{Multi-Agent Collaboration with LLMs.} Recent research extensively explores frameworks enabling effective collaboration among multiple LLM agents, aimed at addressing complex cognitive and decision-making tasks~\cite{li2023camel,zhao2025sirius,estornell2024multidebate1}. 
A prominent paradigm involves mimicking human collaborative dynamics through explicit ``role-playing" mechanisms, where LLM agents are given specialized functions corresponding to organizational roles~\cite{hong2024metagpt}, while~\citet{Chen2023AgentVerseFM} explores multi-agent collaboration via prompting-based interactions.  
Alternative frameworks further enrich multi-agent collaboration through voting and consensus mechanisms~\cite{wang2023SC,park2025maporl,li2024moreNeed}, collective reasoning or discussion-based methodologies~\cite{chen2024reconcile}, and structured agentic debate approaches~\cite{du2024multiagentDebate,liang2024encouragingDebate}, aiming at enhancing factual accuracy and logical consistency. 
Prevalent multi-LLM collaboration frameworks lack theoretical grounding and offer no guarantees of convergence, stability, or cooperation. Our MAC-SPGG framework introduces PGG-inspired incentives to enable collaboration via utility-aligned rewards and structured inter-agent reasoning.

\noindent\textbf{Mechanism Design and Game Theory in LLMs.} Integrating mechanism design and game-theoretic insights into multi-agent LLM systems is increasingly investigated. 

LLM's rationality has been primarily tested. \citet{mao2024alympics} rigorously evaluated LLM strategic behaviors across game-theoretic scenarios, while~\citet{pan2025InfinitelyGames} showed that Bayesian reasoning frameworks encourage cooperative strategies in repeated games among LLM agents, demonstrating cooperative behaviors under suitable incentives in structured games like Public Goods Games (PGGs)~\cite{sreedhar2025simulatingCooperative}. 
Recent work further introduces structured game-theoretic workflows
to improve LLMs’ strategic rationality in both complete- and incomplete-information games~\cite{Hua2024Gametheoretical}. 
Some empirical studies also indicate LLMs exhibit rational behaviors in strategic settings, emphasizing historical context in shaping interactions~\cite{zhang2024LLMrational,akata2025repeatedgameswithLLM,brookins2024playingwithGPT,lore2023strategic}. 

While prior work uses game-theoretic tasks to evaluate LLM rationality, the integration of game and LLM has not been investigated thoroughly. Recent studies have developed tailored incentive mechanisms, such as token auctions, promoting collaboration among agents~\cite{dutting2024mechanismdesign}. 
\citet{cheng2024selfplaying} embedded games to enhance the intrinsic reasoning capabilities of LLMs, demonstrating significant performance improvements across various reasoning benchmarks. Methods like multi-stakeholder alignment significantly enhance LLM output alignment in value-conflict environments~\cite{sel2024skin}.

We propose a new multi-agent collaboration framework grounded in the strategic structure of the SPGG, which demonstrates strong empirical and theoretical effectiveness across diverse tasks.


\begin{figure*}[htbp]
    \centering
    \includegraphics[width=0.61\textwidth]{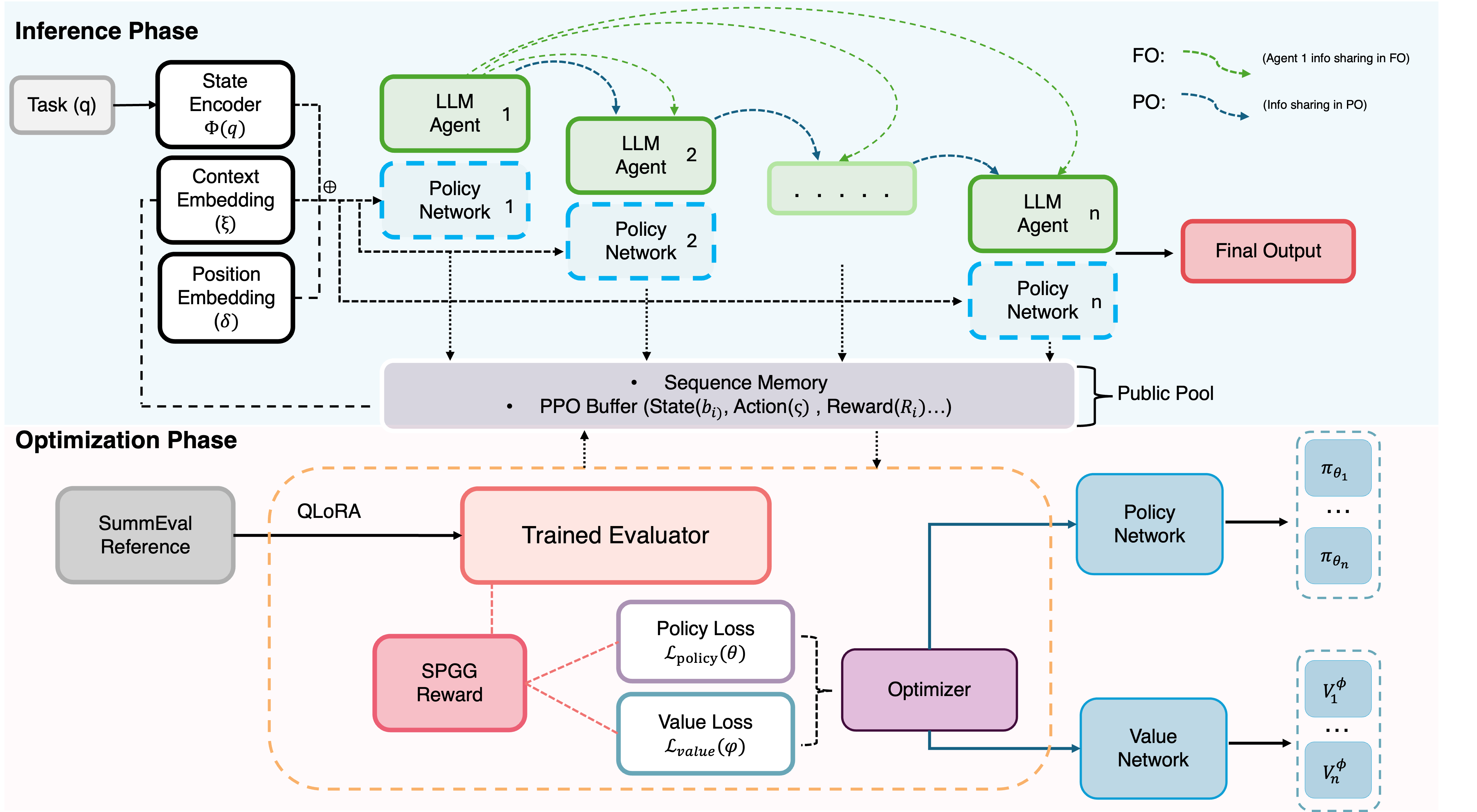}
    \caption{MAC-SPGG Framework. Top: \textit{The Inference Phase}, where LLM agents act in sequence, conditioned on (Partial/Full) observation regimes. Bottom: \textit{The Optimization Phase}, where SPGG rewards drive PPO updates for policy and value networks.}
    \label{fig:spgg_workflow}
\end{figure*}

\section{Method}
\label{sec:spgg}

In this section, we first introduce the fundamental formulation of our MAC-SPGG design, the inference phase in our framework. We then propose the crucial reward structure, followed by the theoretical guarantee of MAC-SPGG. Lastly, we describe the MAC-SPGG learning framework, the optimization phase; see Figure~\ref{fig:spgg_workflow}. The training process is concluded in Algorithm~\ref{alg:spgg_training}, and a comprehensive notation table is summarized in Appendix~\ref{app-notations}.



\subsection{MAC-SPGG Formulation}

To model multi-agent collaboration among $n$ LLM agents performing a shared textual task $q$, we assume the coopetition process follows a finite-horizon, sequential, and decentralized setting. 
Each agent $i$ sequentially provides exactly one \emph{contribution} $\tau_i$ toward the final collective outcome,
\begin{equation}\label{tau-def}
    \tau_i = T_i(h_i, q).
\end{equation}
Here, the function $T_i$ represents the LLM base model of agent $i$, while $h_i$ represents the historical information that is observable to agent $i$. We name the observable history and task information $(h_i, q)$ as \emph{local knowledge}, where all participants make their own contributions based on it. 

For the history $h_i$, we have two modes of observations under the MAC-SPGG framework: (1) {\bf Partial Observation (PO):} The agent $i$ can observe only the contribution from the immediately preceding agent (if any), $h_i^{PO} = \{\tau_{i-1}\}$, and (2) {\bf Full Observation (FO):} The agent $i$ can observe all contributions made by previous agents, $h_i^{FO} = \{\tau_1, \tau_2, \dots, \tau_{i-1}\}$.

In the PO schema, agent $i$ only observes the immediate predecessor’s contribution $\tau_{i-1}$, following the SPGG~\cite{Anwar2023PositionUI, Gallice2018CooperationIS} setting, which is similar to the sense of Markov decision process. In contrast, the agents under the FO regime have full access to the complete history of prior contributions. Both types of observation settings exist in multi-agent LLM studies~\cite{du2024multiagentDebate, Wu2023AutoGenEN}. Although the coordinator-free mechanism of MAC-SPGG saves computation resources, the FO mode would consume more tokens than the PO mode. Such a difference in information availability and resource usage will lead to distinct comprehensibility in various types of tasks in our experiment.


\begin{remark}[No-Observation Regime]
\normalfont
When $h_i=\emptyset$, agents have no cross-agent observability and act only on the task context $q$. This reduces to the simultaneous-move PGG setting~\cite{suurmond2004reputation,Andreoni1988WhyFR}. Among the existing multi-agent LLM frameworks, ECON~\cite{ECON} is the closest analogue: a central coordinator prescribes strategies to otherwise non-communicating agents. 
Because MAC-SPGG is coordinator-free and sequentially observable, the No-Observation regime is incompatible with the model. We omit the No-Observation setting and retain ECON as an experimental benchmark.
\end{remark}

After all agents have committed their contribution, the contribution $\tau_i$ of each agent will be evaluated by a task-specific metric, the \emph{score}, $c_i(\tau_i, q)$, and a model-related metric, the \emph{cost}, $\ell_i(\tau_i, q, T_i)$. 
The score indicates the performance of the contribution, which is evaluated by a given task-specific function ${\mathcal E}$, $c_i={\mathcal E}(\tau_i, q)$. For instance, in multiple-choice tasks, the score represents the accuracy of the test; in more complex tasks, such as a generation task, the score is evaluated by a fine-tuned evaluator; see training details in Appendix D. We denote the score by $c_i(\tau_i, q)$ to show its relevance to $\tau_i$ and $q$.
For the cost part, under the usage of LLM, the number of consumed tokens would be a straightforward measure of cost, and different base models $T$ will lead to various levels of token usage.

We denote the final task score used to assess success by $C(\vec{\tau}, q)$, defined as the last agent’s contribution:
\begin{equation}
    C(\vec{\tau}, q) = c_n(\tau_n, q).
\end{equation}

The task $q$ succeeds if the final score $C(\vec{\tau},q)$ surpasses a predefined threshold $B(q)$.
Our objective is to maximize the final score $C$ rather than merely exceed the threshold $B$ on task $q$ by efficiently utilizing LLM agents to collaborate on the shared task.

\begin{remark}[Cumulative Effect]
\normalfont Although other agents' contribution is not on the surface of $c_n(\tau_n, q)$, we still denote the final score $C$ as a function of all the contributions $\vec{\tau}$ due to the cumulative effect of the MAC-SPGG. Different from PGG, where the final performance is calculated by summing up all the contributions, the nature of multi-agent LLM tasks and prompting needs a summary step instead of concatenating the AI-generated content (AIGC) directly. In ECON~\cite{ECON} or other coordinator-based frameworks, a summary agent in the last step would absorb all the others' outputs and generate the final answer. In our MAC-SPGG framework, predecessors' outputs have already been embedded into the sequential process. For instance, if we are under the FO mode, where $c_n=T_n(h_n,q)$, $h_n$ contains all the previous $\tau_i$ information. If we are under the PO mode, we can regard the final score as
\[
\begin{aligned}
    C(\vec{\tau}, q) = &c_n\left(T_n(\tau_{n-1},q), q\right)\\
    = & c_n\left(T_n\left(T_{n-1}(\tau_{n-2},q),q\right), q\right)\quad\cdots \\
    = & c_n\left(T_n\left(T_{n-1}(\cdots \left(T_1(q),q\right)\cdots,q\right), q\right).
\end{aligned}
\]
In such a context, the impact of each contribution $\tau_i$ on the final score is not explicit, but in an iterative way.
\end{remark}

\subsection{Reward Structure and Equilibrium}
The basic form of the MAC-SPGG structure enables collaboration among LLM agents, but it is not guaranteed to be efficient. As traditional PGG-related research has revealed, the equilibrium may collapse into a situation where no one contributes, and hence the whole task fails~\cite{Ledyard1994PublicGA}. Although the MAC-SPGG framework has been found valuable through both theoretical and experimental analyses~\cite{Anwar2023PositionUI, Gallice2018CooperationIS}, the benefits of involving a sequential decision process remain unclear. We believe more can be done with the LLM multi-agent collaboration tasks.

In our model, unlike the traditional utility in PGG or SPGG, we develop a special \emph{synergy-aligned reward} structure to stimulate the LLM agent's contribution. The reward function, $R$, is known to the agents, but the actual reward will be revealed only after the final judgment is made. 


\begin{defi}[Synergy-aligned Reward]\label{reward-def} The reward for agent \( i \in \{1, \dots, n\} \) is defined as:
\[
\begin{aligned}
    R_i = &-l_i(\tau_i,q,T_i) + \gamma \cdot \frac{c_{i}(\tau_i,q)}{B(q)} \cdot c_i(\tau_i,q)\\
    &\quad +\frac{\rho}{n} \cdot C(\vec{\tau},q) - P \cdot {\bf 1} \left( C(\vec{\tau},q) < B(q) \right).
\end{aligned}
\]
\end{defi}

Here, the first row represents the measurement of the current status at decision time based on the possibly observable history, while the last row captures the endgame status and the outcome of the task.
Besides the individual cost $l_i$, three hyper-parameters are involved to promote alignment between individual incentives and task-level success: (i) task reward multiplier \( \rho > 0 \), which controls the magnitude of global utility sharing and encourages agents to work toward collective success; (ii) cooperation coefficient \( \gamma > 0 \) for history-aware cooperation bonus, which scales the intermediate reward based on accumulated contribution, promoting alignment with sequential synergy; and (iii) failure penalty \( P > 0 \) balances collective penalty, which discourages free-riding by penalizing all agents if the task fails. 

As discussed in Section~\ref{sec:literature}, though LLM agents are not inherently utility-maximizing, recent work shows they exhibit quasi-rational behavior under appropriate conditioning. In MAC-SPGG, rationality is induced through prompt engineering and reward-based training. To fill in the gap between LLM and traditional game-theoretical analysis, we make the following two assumptions about the LLM action space.

\begin{assum}[Score Assumption]\label{score-assum}
The score $c_i$ of each agent \( i \in \{1, \dots, n\} \) is positive, bounded, and finite
\[ c_i\in [c_{\min}, c_{\max}], \quad \mbox{ where} \quad 0 < c_{\min} \leq c_{\max} < \infty. \]
The upper bound \( c_{\max} \), which defined by
\[
c_{\max} \equiv \sup_{\vec\tau} \big\{ c_n\big( \tau_n(\tau_{n-1}(\cdots (\tau_1(q),q) \cdots),q) ,q\big) \big\},
\]
can surpass the task-specific threshold $c_{\max}\geq B(q)$
\end{assum}
The positive lower bound reflects the empirical observation that LLMs typically produce non-trivial outputs when prompted, ensuring a minimum contribution from each agent. It may imply that the last agent can complete the task when recursively conditioned on upstream outputs. 
\begin{assum}[Cost Assumption]\label{cost-assum}
The individual cost function \( \ell_i(c_i) \) is strictly convex and twice continuously differentiable over \( [c_{\min}, c_{\max}] \), and \( \ell_i'(c_i) > 0 \).
\end{assum}
The cost assumption follows a naive belief that higher-quality outputs often require longer sequences and greater inference resources with an increasing marginal cost~\cite{Chowdhery2022PaLMSL,Kaplan2020ScalingLF}. Given the mathematical support from two assumptions, we have
\begin{theorem}[Equilibrium]\label{spne}
Under a reasonable cooperation coefficient $\gamma$ and failure penalty $P$, where
\[
\begin{aligned}
\rho &> n\cdot\max_{i}\ell_i'(c_{\max}),\\
\gamma &> \max_{k=2,\dots,n}
         \frac{\ell_k'(c_{\max})\cdot B(q)-\rho/n}{c_{\min}/B(q)}, \mbox{ and }\\
P &> \Bigl(\max_i\left\{ \ell_i'(c_{\max})\right\}+\gamma\tfrac{c_{\max}}{B(q)}
            +\tfrac{\rho}{n}\Bigr)\cdot(c_{\max}-c_{\min}),
\end{aligned}
\]
there exists a joint strategy \( \mathbf{c}_i^* = (c_1^*, \dots, c_n^*)\) that constitutes a \textit{\textbf{unique}} Subgame Perfect Nash Equilibrium (SPNE),
\[
\mathbf{c}_i^* \in \arg\max_{\vec{c}} \left\{ \text{SPNE under } R_i \right\},
\]
where every agent $ i\in \{1, \dots, n\}$ contributes positively, \( c_i^* > 0 \), and the overall task would succeed \( C(\vec{\tau},q) \geq B(q) \).
\end{theorem}
Theorem~\ref{spne} shows the existence and uniqueness of the SPNE under our MAC-SPGG framework, which enables the rationality of LLM agents. Under SPNE, each agent contributes positively to the cooperation, escaping from the ``bad'' free-riding equilibrium in PGG. Our equilibrium results from not only the stimulated reward function, but also Assumption~\ref{score-assum}. LLM agents must contribute, while a rational human may dedicate no effort to the task. This feature also leads to a comparative static analysis.
\begin{theorem}[Comparative Statics Analysis]\label{comp-stat}
    Under the MAC-SPGG equilibrium, total welfare increases with both the cooperation incentive $\gamma$ and the public-good sharing rate $\rho$, but decreases with the task threshold $B$. 
\end{theorem}
These monotonic relationships hold as long as all LLM agents contribute non-negatively in equilibrium. Detailed proofs of Theorems~\ref{spne} and~\ref{comp-stat} appear in Appendix~\ref{app-proof}, and numerical verification is provided in Appendix~\ref{app-spne}.

\begin{algorithm}[tb]
\caption{MAC-SPGG Framework}
\label{alg:spgg_training}
\begin{algorithmic}[1]
\Require Initial prompt $q$; base models $\{T_i\}_{i=1}^n$; evaluator $\mathcal{E}$; game parameters $\rho, \gamma, P, B(q)$; max episodes $T_{\text{max}}$; early stopping thresholds $R_{\text{th}}, C_{\text{target}}, \epsilon$
\Ensure Optimized policy and value function parameters $\{\theta_i^*, \phi_i^*\}_{i=1}^n$

\State Initialize $\{\theta_i, \phi_i\}_{i=1}^n$, encoder $\theta_{\Phi}$, buffer $\mathcal{D}$, and history $\mathcal{H}$

\For{episode $t = 1$ to $T_{\text{max}}$}
    \State Reset $\mathcal{D} \gets \emptyset$, $\mathcal{H} \gets \emptyset$
    
    \For{agent $i = 1$ to $n$} \Comment{Sequential rollout}
        \State Extract task embedding $\Phi(q)$, context features $\xi_i$ and position embedding $\delta_i$
        \State Construct $b_i \gets [\Phi(q); \xi_i; \delta_i]$
       
        \State Sample config $\vec{\varsigma}_i \sim \pi_{\theta_i}(\cdot \mid b_i)$
        \State Generate output $\tau_i \gets T_i(q, h_i|\vec{\varsigma}_i)$ 
        \State Store $(b_i, \vec{\varsigma}_i, \tau_i)$ in $\mathcal{D}$, update $\mathcal{H} \gets \mathcal{H} \oplus \tau_i$
    \EndFor

    \For{agent $i = 1$ to $n$} \Comment{Reward computation}
        \State Evaluate quality $c_i \gets \mathcal{E}(\tau_i, q)$
        \State Compute reward $R_i$, advantage $A_i = R_i - V^{\phi_i}(b_i)$
        \State Store $(R_i, A_i)$ in $\mathcal{D}$
    \EndFor

    \State Compute $\bar{R}_t = \frac{1}{n} \sum_{i=1}^n R_i$, $\bar{C}_t = \frac{1}{n} \sum_{i=1}^n c_i$

    \For{agent $i = 1$ to $n$} \Comment{PPO update}
        \State Update $\theta_i$, $\phi_i$ via gradient descent on $\mathcal{L}_{\text{PPO}}$
        
    \EndFor

    \If{$\bar{R}_t \ge R_{\text{th}}$ and $\bar{C}_t \ge C_{\text{target}}$ and\\
    \hspace{1em} $|\bar{R}_t - \bar{R}_{t-1}| \le \epsilon$ and $|\bar{C}_t - \bar{C}_{t-1}| \le \epsilon$}
        \State \textbf{break} \Comment{Early stopping}
    \EndIf
    
\EndFor

\State \textbf{return} $\{\theta_i^*, \phi_i^*\}_{i=1}^n$
\end{algorithmic}
\end{algorithm}

\subsection{RL as a Meta-Control Framework}
\label{section:rl_meta_control}

To operationalize the theoretical framework, we conceptualize each agent's generation function \( \mathcal{G}_i \) as a two-phase process shown in Figure~\ref{fig:spgg_workflow}. At the \textit{Inference Phase}, a foundational language model \( {T}_i \) generates textual outputs from the MAC-SPGG mechanism, while at the \textit{Optimization Phase}, an RL-trained \emph{meta-policy} \( \pi_{\theta_i}\) is trained to synthesize strategic configurations from high-level belief representations, enabling adaptive and coordinated contributions across agents. We employ independent PPO learners~\cite{PPO} for each agent under the synergy-aligned reward structure.

The generation process for agent \( i \) is cast as a hierarchical control problem given a prompt of the task \( q \). First, the agent constructs an enhanced belief state vector \( b_i = [\Phi(q); \xi_i; \delta_i] \) by concatenating a task embedding \( \Phi(q) \), context features \(\xi_i \) containing historical performance and environmental information, and a position embedding \( \delta_i \). This belief \( b_i \) informs the agent's meta-policy \( \pi_{\theta_i} \), which generates a generative configuration vector, \( \vec{\varsigma}_i \sim \pi_{\theta_i}(\cdot | b_i) \). As a result, this vector serves as a local policy to direct the global collaboration. Finally, the LLM produces the agent's contribution \( \tau_i \) as Eq.~(\ref{tau-def}) under this strategic guidance $T_i (q, h_i| \vec{\varsigma}_i)$. 

The verification-based reward formulation in Definition~\ref{reward-def} enables structured feedback under sequential collaboration. Once agent \( i \) makes a strategic decision based on its belief \( b_i \), the policy optimization will handle the observable part, while the value approximation deals with the rest.


We train each agent’s meta-policy \( \pi_{\theta_i} \) using a decentralized actor-critic method based on PPO. Each agent operates in a one-step decision process per episode, observing its belief state \( b_i \), sampling a continuous configuration vector \( \vec{\varsigma}_i \), and generating a textual contribution via its base LLM.

The value function over belief states is defined as:
\begin{equation}
    V_i^{\phi}(b_i) = \mathbb{E}_{\pi_{\theta_i}}[R_i | b_i],
\end{equation}
where \( R_i \) is the total episodic reward in Definition~\ref{reward-def}. We estimate the advantage using Generalized Advantage Estimation (GAE)~\cite{schulman2016gae} over single-step rollout:
\begin{equation}
A(b_i, \vec{\varsigma}_i) = R_i + \gamma \cdot V_i^{\phi}(b_{i+1}) - V_i^{\phi}(b_i),
\end{equation}
where \( V_i^{\phi}(b_{i+1}) \) is the terminal value, typically set to zero under the one-step assumption. Hence, each agent’s PPO objective can be defined as:
\begin{equation}\label{eq:ppo_loss}
\begin{aligned}
\mathcal{L}_{\text{PPO}}&(\theta_i) = - \lambda_{\text{value}} \cdot \big( V_i^{\phi}(b_i) - R_i \big)^2\\
& +\mathbb{E}_{b_i, \vec{\varsigma}_i} \Big[ \min \big( R(\theta_i) \cdot A(b_i, \vec{\varsigma}_i), \text{clip}_\varepsilon(R(\theta_i)) \cdot A(b_i, \vec{\varsigma}_i) \big) \Big], \\
\end{aligned}
\end{equation}
where the importance sampling ratio is:
\begin{equation}
    R(\theta_i) = \pi_{\theta_i}(\vec{\varsigma}_i \mid b_i)/\pi_{\theta_{\text{old}},i}(\vec{\varsigma}_i \mid b_i),
\end{equation}
where $R(\theta_i) $ represents the ratio between current and previous policies, $A(b_i, \vec{\varsigma}_i)$ is the estimated advantage, and $V_i^{\phi}(b_i)$ is the learned value function.
The coefficient $\lambda_{\text{value}}$ weights the contribution of the value loss in the overall objective. 


\begin{table*}[t]
\centering
\small
\begin{tabular}{llcccccc}
\toprule\midrule
\textbf{System Category} & \textbf{Configuration} & \textbf{\#Params} & \textbf{HumanEval} & \textbf{MMLU} & \textbf{GSM8K} & \textbf{SummEval (Avg)} \\
\midrule
\multirow{3}{*}{Zero-Shot COT Single-Agent}
& SmolLM2-1.7B-Instruct & 1.7 & 24.4 (-49.38) & 29 (-46) & 45 (-50) & 4.607 (-0.12) \\
& Llama3.1-8B-Instruct & 8 & 59.76 (-14.02) & 57 (-18) & 88 (-7) & 4.638 (-0.09) \\
& Qwen3-8B & 8 & 64.63 (-9.15) & 66 (-9) & 89 (-6) & 4.677 (-0.05) \\
\midrule
\multirow{3}{*}{Few-Shot COT Single-Agent}
& SmolLM2-1.7B & 1.7 & 29.9 (-43.88) & 41 (-34) & 52 (-43) & -- \\
& Llama3.1-8B & 8 & 72.6 (-1.18) & 70 (-5) & 90 (-5) & -- \\
& Qwen3-8B & 8 & 72.0 (-1.78) & 67 (-8) & 92 (-3) & -- \\
\midrule
\multirow{4}{*}{Multi-Agent Baselines}
& Majority Voting & 17.7 & -- & 71 (-4) & 84 (-11) & -- \\
& Multi-Agent Debate & 17.7 & -- & 66 (-9) & 86 (-9) & -- \\
& CAMEL & 16 & 48.78 (-24.99) & 42 (-33) & 88 (-7) & -- \\
& ECON & 25.7 & 70.73 (-3.05) & 64 (-11) & 89 (-6) & 4.590 (-0.14) \\
\midrule
\multirow{2}{*}{MAC-SPGG Framework (Ours)}
& MAC-SPGG (PO) & 17.7 & 67.07 (-6.71) & \textbf{75 (-)} & \textbf{95 (-)} & 4.449 (-0.28) \\
& MAC-SPGG (FO) & 17.7 & \textbf{73.78 (-)} & 69 (-6) & 93 (-2) & \textbf{4.728 (-)} \\
\bottomrule
\end{tabular}
\caption*{\footnotesize\textit{Note.} ``-'' indicates not applicable, e.g., voting-based methods cannot generate coherent outputs for HumanEval or SummEval}
\caption{Performance on four benchmarks with delta (in parentheses) relative to the best MAC-SPGG setup. Adopted metrics: HumanEval in Pass@1 (\%), MMLU and GSM8K in accuracy (\%), and SummEval in the averaged human evaluation (0--5).}
\label{tab:main_results}
\end{table*}

To ensure efficient optimization and convergence, we apply an early stopping mechanism based on the empirical performance. Specifically, training is terminated once two external criteria are jointly satisfied. First, the average episodic reward across agents exceeds a predefined threshold, \( \sum_{i=1}^n R_i/n \geq R_{\text{threshold}} \). Second, the average evaluator-assessed output quality meets or surpasses a target value, \( \bar{C} \geq C_{\text{target}} \). Here, \( \bar{C} \) denotes the average of final task scores \( C(\vec{\tau}, q) \) across evaluation episodes.
Also, we monitor convergence stability by requiring both the average reward and quality scores to remain within a small margin \( \epsilon \) across consecutive episodes, \( |\bar{R}_{t+1} - \bar{R}_t| \leq \epsilon \) and \( |\bar{C}_{t+1} - \bar{C}_t| \leq \epsilon \), to ensure training halts only after meaningful improvements have plateaued. This early stopping strategy ensures that agents not only achieve high collaborative performance but also maintain consistent quality in generation.

\section{Experiment}\label{sec:exp}
This section outlines the experimental setup, reports effectiveness performance comparisons with various benchmarks, sequential ordering effect analysis, and presents ablation studies on base model combinations and heterogeneity.

\subsection{Datasets}





We evaluate our workflow on four standard benchmarks spanning diverse capabilities: \textit{HumanEval}~\cite{chen2021humanEval} for code generation (Python tasks with unit-test evaluation), \textit{MMLU}~\cite{hendrycks2020mmlu} for general knowledge and reasoning (57 subjects across STEM and humanities), \textit{GSM8K}~\cite{cobbe2021gsm8k} for multi-step arithmetic problem solving (grade-school math word problems), and \textit{SummEval}~\cite{fabbri2021summeval} for natural language understanding (human-annotated summaries rated on coherence, consistency, fluency, and relevance). For \textit{SummEval}, we train a reinforcement learning-based evaluator aligned with human-centric metrics; see Appendix~\ref{app-exp-summeval}.

\subsection{Comparison Methods}

We compare MAC-SPGG against several widely adopted strong baselines: (1) \textit{\textbf{Zero-shot} CoT prompting}~\cite{Kojima2022zeroShot}:  
Directly asks the model to reason step-by-step without any examples.
(2) \textit{\textbf{Few-shot} CoT prompting}~\cite{Wei2022COT}:  
Provides a few worked-out examples to guide the model’s step-by-step reasoning. (3) \textit{\textbf{Majority Voting}-based multi-agent ensemble} ~\cite{li2024moreNeed}: Multiple independent agents generate answers in parallel, and the final output is selected via majority vote or other aggregation strategies. (4) \textit{\textbf{Multi-Agent Debate}-style prompting}~\cite{du2024multiagentDebate}: Agents engage in argumentation or critique each other's outputs before converging on a final decision. (5) \textit{\textbf{CAMEL}-style role-based collaboration}~\cite{li2023camel}: Agents are assigned distinct roles (e.g., user, assistant, critic) to simulate structured dialogues. (6) \textit{\textbf{ECON}}~\cite{ECON}: Agents act independently without observing each other, controlled and manipulated by one coordinator. 

\subsection{MAC-SPGG Setups}

In our experiments, we instantiate the MAC-SPGG framework using three sequentially interacting language models. The full set of training hyperparameters is provided in Appendix~\ref{app-exp-rl}. We focus primarily on training and evaluating \textit{small-scale language models} under the MAC-SPGG setting. As heterogeneous model integration has been shown to enhance multi-agent reasoning and strategic capabilities~\cite{park2025maporl,xu2025multiagentft}, we specifically employ Qwen3-8B~\cite{qwen3-8B}, SmolLM2-1.7B~\cite{smollm2}, and LLaMA 3.1-8B~\cite{llama3.1-8B} to exploit model heterogeneity effectively.


\subsection{Main Results}

We show the performance of each method across four representative evaluation tasks 
in Table~\ref{tab:main_results}. The MAC-SPGG, under both PO and FO regimes, consistently outperforms most single-agent and multi-agent baselines, particularly excelling on complex reasoning tasks such as GSM8K and MMLU. To provide reference points for upper-bound performance, we include GPT-3.5 Turbo~\cite{GPT3.5Turbo}, GPT-4-0613~\cite{ChatGPT4}, and Qwen2.5-72B-Instruct~\cite{Qwen3-72B-Instruct} in a zero-shot setting, without fine-tuning. We find that our MAC-SPGG achieves competitive performance with significantly fewer total parameters. Details could be found in Appendix~\ref{app-exp-large}. These results highlight the effectiveness of our cooperative mechanism in MAC-SPGG: by strategically leveraging multiple smaller models and incentivizing collaboration through game-theoretic design, our framework achieves strong performance with substantially fewer parameters. For a detailed case study, we refer readers to Appendix~\ref{app-case}.

\subsection{Agent Sequential Ordering Effects}




From Table~\ref{tab:agent_order_ablation}, we observe three insights: (i) \emph{Sequencing matters}: under PO, \texttt{LLaMA $\rightarrow$ Smol $\rightarrow$ Qwen} attains the highest MMLU accuracy (78\%), while \texttt{Smol $\rightarrow$ LLaMA $\rightarrow$ Qwen} leads on GSM8K (95\%), indicating task-dependent optima shaped by task complexity and model capabilities. (ii) \emph{Avoid ``poor'' summarizer}: performance often degrades when ending with the smallest model, as the last agent bears greater responsibility in cumulative decision-making and, under PO, has limited backward correction, constraining its ability to refine complex outputs. (iii) \emph{More context is not always better}: FO’s full access does not guarantee superior results, as PO can outperform FO when excess information introduces redundancy or distractions, echoing a ``less is more'' effect. Together, these findings highlight the nuanced effects of agent ordering and offer actionable guidance for multi-agent design.

\begin{table}[htbp]
\fontsize{9pt}{9pt}\selectfont
\small
\centering
\setlength{\tabcolsep}{3.5pt} 
\begin{tabular}{llcc}
\toprule\midrule
\textbf{Setting} & \textbf{Agent Order} & \textbf{MMLU} & \textbf{GSM8K} \\
\midrule
\multirow{6}{*}{PO}  
& Qwen → LLaMA → Smol & 56 & 66 \\
& Qwen → Smol → LLaMA & 74 & 91 \\
& Smol → Qwen → LLaMA & 76 & 91 \\
& LLaMA → Smol → Qwen & \textbf{78} & 93 \\
& LLaMA → Qwen → Smol & 48 & 71 \\
& Smol → LLaMA → Qwen & 75 & \textbf{95} \\
\midrule
\multirow{6}{*}{FO}  
& Qwen → LLaMA → Smol & 49 & 61 \\
& Qwen → Smol → LLaMA & \textbf{77} & 90 \\
& Smol → Qwen → LLaMA & 76 & 90 \\
& LLaMA → Smol → Qwen & 72 & \textbf{96} \\
& LLaMA → Qwen → Smol & 44 & 72 \\
& Smol → LLaMA → Qwen & 69 & 93 \\
\bottomrule
\end{tabular}
\caption{Ablation Study of Agent Ordering under Partial Observation (PO) and Full Observation (FO) settings.}\label{tab:agent_order_ablation}
\end{table}

\subsection{Efficiency Analysis}
\begin{figure}[htbp]
    \centering
    \includegraphics[width=1.05\linewidth]{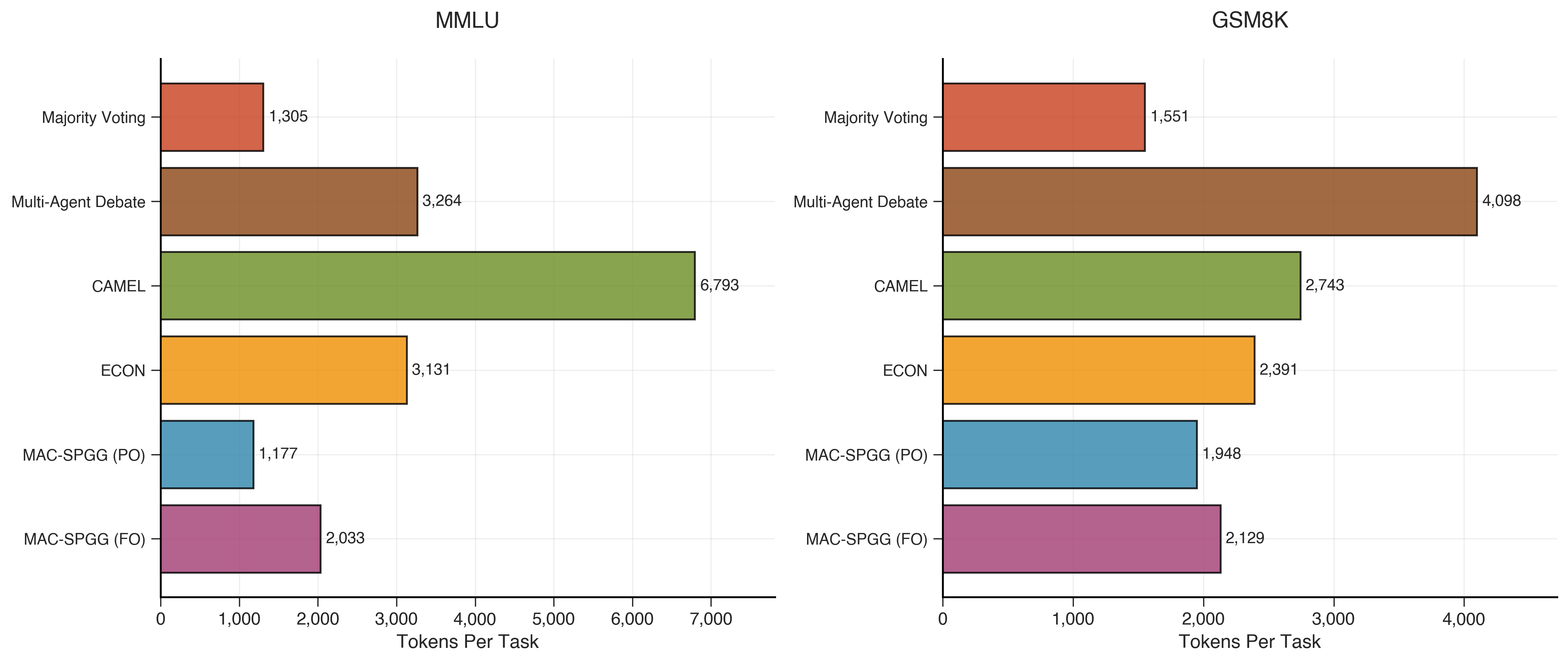}
    \vspace{-2ex}
    \caption{Token usage per task across different collaboration frameworks. MAC-SPGG significantly reduces token consumption under both Full and Partial observation settings.}
    \label{fig:tokens}
    \vspace{-2ex}
\end{figure}

We also conducted a cost efficiency analysis by comparing the token usage per task across different collaboration frameworks, as shown in Figure~\ref{fig:tokens}. The results indicate that MAC-SPGG consistently achieves lower token consumption in both PO and FO settings compared to other baselines.
Specifically, the MAC-SPGG mechanism under PO achieves the lowest token usage, demonstrating significant efficiency gains. This reduction in tokens highlights the economic advantage of MAC-SPGG, as it effectively leverages structured collaboration, minimizing communication overhead while maintaining or improving task performance.

\subsection{Ablation Study}
\label{Eep:Ablation}

To understand the effectiveness of the MAC-SPGG mechanism and the role of agent heterogeneity, we conducted an ablation study presented in Table \ref{tab:spgg_ablation}. 
First, enabling the MAC-SPGG mechanism consistently improves performance across both PO and FO settings, which highlights the efficacy of our framework.
Second, we chose to employ three Qwen (\textit{Qwen×3}) models in our experiments due to their consistently superior performance across all evaluated benchmarks. Using the strongest available model ensures that our observed results accurately reflect the capabilities and potential benefits of the MAC-SPGG framework, without confounding factors introduced by model heterogeneity.

Overall, these findings emphasize that both the MAC-SPGG mechanism and agent heterogeneity are essential to task performance, 
which should be carefully balanced when designing multi-agent cooperative systems.

\begin{table}[htbp]
\centering
\small
\setlength{\tabcolsep}{3pt}
\begin{tabular}{llcccc}
\toprule\midrule
\textbf{Obs.} & \textbf{Agents} & \textbf{Het.} & \textbf{SPGG} & \textbf{MMLU} & \textbf{GSM8K} \\
\midrule
\multirow{4}{*}{PO} 
& LLaMA + Smol + Qwen & \checkmark & \checkmark & \textbf{78} & 93 \\
& LLaMA + Smol + Qwen & \checkmark &     & 72 & 79 \\
& Qwen×3              &       & \checkmark & \textbf{78} & \textbf{94} \\
& Qwen×3              &      &     & 71 & 77 \\
\midrule
\multirow{4}{*}{FO} 
& LLaMA + Smol + Qwen & \checkmark & \checkmark & 72 & \textbf{96} \\
& LLaMA + Smol + Qwen & \checkmark &     & 71 & 77 \\
& Qwen×3              &      & \checkmark & \textbf{80} & 95 \\
& Qwen×3              &     &      & 68 & 74 \\
\bottomrule
\end{tabular}
\caption{Ablation study on MAC-SPGG mechanism and agent heterogeneity (accuracy \%).}\label{tab:spgg_ablation}
\end{table}

\section{Conclusion}
This paper presents a principled framework for structured cooperation among LLM-based agents, grounded in the theory of Sequential Public Goods Games (SPGG). By embedding incentive-compatible mechanisms into the agent interaction protocol, our approach induces conditional cooperation, belief propagation, and sequential adaptation—capabilities rarely addressed in existing multi-agent LLM systems. Through extensive empirical evaluations, we show that MAC-SPGG not only improves performance across diverse tasks and observation regimes but also enhances cost efficiency by minimizing redundant communication.

More broadly, this work advances the methodological foundation for aligning autonomous language agents through economic incentives and strategic reasoning. Rather than relying on ad-hoc coordination heuristics or static voting rules, MAC-SPGG formalizes collaboration as a dynamic process shaped by information flow and strategic interdependence. The observed benefits of SPGG in both full and partial observability regimes suggest a promising direction: treating cooperation not as an engineered protocol, but as an emergent equilibrium behavior shaped by incentives.

Our findings invite further exploration into mechanism design for large-scale multi-agent LLM systems, particularly in settings involving partial knowledge, bounded rationality, or open-ended objectives. We believe this work takes an essential step toward scalable, mechanism-grounded, and adaptive cooperation among foundation models.

\bibliography{main}


\clearpage 

\noindent{\bf\Large Appendices of \textit{Everyone Contributes! Incentivizing Strategic Cooperation in Multi-LLM Systems via Sequential Public Goods Games}}
\appendix
\numberwithin{equation}{section}
\numberwithin{figure}{section}
\numberwithin{table}{section}

\section{Notation}
\label{app-notations}

This section summarizes the notations used throughout the paper, categorized for clarity.

\begin{table*}[htbp]
\centering
\fontsize{9pt}{11pt}\selectfont
\begin{tabular}{p{0.15\textwidth}p{0.32\textwidth} p{0.15\textwidth}p{0.32\textwidth}}
\toprule\midrule
\textbf{Symbol} & \textbf{Meaning} & \textbf{Symbol} & \textbf{Meaning} \\
\midrule

\multicolumn{4}{c}{\textbf{General Notations}} \\\midrule
$n$ & Total number of agents in the system & $q$ & The shared task \\
$i, k, j$ & Index for a specific agent & $\tau_i$ & The contribution text from agent $i$ \\
$T_i$ & The base Large Language Model (LLM) for agent $i$ & $\vec{\tau}$ & The vector of all agents' contributions \\
$h_i$ & The observable history available to agent $i$ & $h_i^\text{PO}$ & History under Partial Observation \\
$h_i^{\text{FO}}$ & History under Full Observation & $\mathcal{G}_i$ & The generation function of agent $i$ \\ $T_{\text{max}}$ & Maximum number of training episodes \\

\midrule
\multicolumn{4}{c}{\textbf{Reinforcement Learning (RL) Framework}} \\\midrule
$s_t$ & State vector for the RL agent at step $t$ & $b_i$ & The belief state of agent $i$ \\
$\pi_{\theta_i}$ & The meta-policy of agent $i$ parameterized by $\theta$ & $V_i^{\phi}(b_i)$ & The value function parameterized by $\phi$ \\
$\vec{\varsigma}_i$ & Configuration vector produced by the policy $\pi_{\theta_i}$ & $A(b_i, \vec{\varsigma}_i)$ & The advantage function \\
$\mathcal{L}_{\text{PPO}}$ & The clip-based loss function for PPO & $R(\theta_i)$ & The importance sampling ratio in PPO \\
$\varepsilon$ & The clipping parameter in PPO loss & $\lambda_{\text{value}}$ & The coefficient for the value loss term \\
$\Phi(q)$ & Embedding of the task $q$ & $\xi_i$ & Contextual features (e.g., history) \\
$\delta_i$ & Positional embedding for agent's turn & $\bar{C}, \bar{R}$ & Average score/reward for early stopping \\
$\mathcal{D}$ & Experience buffer & $\mathcal{H}$ & Episode history log \\
$\theta_i^*, \phi_i^*$ & Optimized parameters after training & $R_{\text{th}}, C_{\text{target}}$ & Reward and quality thresholds for early stopping \\
$R_{\text{th}}$ & Experience buffer & $\mathcal{H}$ & Episode history log \\
$r_\text{LoRA}, \alpha, d$ & LoRA training parameters: rank, alpha, and dropout &     $\epsilon$ &  Convergence margin \\
$\bar{R}_t, \bar{C}_t$ & Avg. reward \& quality at episode $t$ & \\

\midrule
\multicolumn{4}{c}{\textbf{MAC-SPGG Mathematical Model}} \\\midrule
$c_i$ & The quality score of an individual contribution $\tau_i$ & $\ell_i$ & The cost associated with generating $\tau_i$ \\
$C(\vec{\tau}, q)$ & The final score of the completed task & $B(q)$ & The predefined threshold for task success \\
$R_i$ & The total reward assigned to agent $i$ & $S_n$ & The cumulative sum of contributions, $\sum c_j$ \\
$\gamma$ & The cooperation coefficient for synergy bonus & $\rho$ & The multiplier for the shared task reward \\
$P$ & The penalty for failing to meet the threshold $B(q)$ & $\mathbf{c}^*$ & The unique Subgame Perfect Nash Equilibrium (SPNE) \\
$W(\cdot)$ & The total welfare function of the system & $\mathbf{1}(\cdot)$ & The indicator function (returns 1 if true, 0 otherwise) \\
$G(\cdot), f(\cdot)$ & Helper functions for payoff analysis in proofs & $\mathcal{A}^+, \mathcal{A}^-$ & Regions and sets for success/failure in proofs \\
$R_n^\bullet, R_n^+, R_n^-$ & Agent's payoff function in different regions & $t_k$ & Minimum contribution for agent $k$ to avoid penalty \\
$c_n^\star, \tilde c_n$ & Optimal and alternative choices in proofs & & \\

\midrule
\multicolumn{4}{c}{\textbf{Evaluator Model}} \\ 
\midrule
$\mathcal{E}(\tau_i, q)$ & Evaluator function that returns the score $c_i$ &   $\mathcal{L}_{\text{eval}}$ & The loss function for training the evaluator model \\
$\mathbf{r}$ & The four-dimensional score vector from SummEval & $r_{\text{...}}$ & Individual scores (relevance, coherence, etc.) \\
$x_i$ & An input document-summary pair for the evaluator & $y_t$ & A target token during evaluator training \\
$\mathcal{T}_\text{score}$ & The set of token indices corresponding to scores & & \\

\bottomrule
\end{tabular}
\caption{Summary of Notations}\label{tab:notation}
\end{table*}

\section{Proof of Theorems~\ref{spne} and~\ref{comp-stat}}\label{app-proof}
\noindent

First, we need to prove a required Lemma.
\begin{lemma}[Monotone Best Response]\label{lemma:monotonic_response}
Under the reward in Definition~\ref{reward-def}, the best‑response contribution
\[
c_i^*(h_i)=c_i\!\bigl(\tau_i^{*},q\bigr)
\]
is \emph{monotonically non‑decreasing} in $c_{i-1}$; that is,
\[
c_{i-1}' > c_{i-1}\quad\Longrightarrow\quad c_i^*(c_{i-1}') \;\ge\; c_i^*(c_{i-1}).
\]
\end{lemma}

\vspace*{.05in}\noindent{\bf Proof of Lemma~\ref{lemma:monotonic_response}: }
We present the argument for the terminal agent $n$; the same reasoning applies to any interior agent $i$ after conditioning on the future best responses.

\noindent\textbf{Step 1:} Rewrite the payoff.
Under Definition~\ref{reward-def}, agent $n$’s payoff is
\begin{align*}
R_n(c_n \mid c_{n-1}) 
&= -\ell_n(c_n) 
  + \gamma \cdot \frac{c_{n-1}}{B(q)} \cdot c_n \\
&\quad + \frac{\rho}{n} \cdot (c_{n-1} + c_n) 
  - P \cdot \mathbf{1}(c_n < B(q)).
\end{align*}

For convenience set
\[
G(c_n,c_{n-1}) =
-\ell_n(c_n)
\;+\;\gamma\,\frac{c_{n-1}}{B(q)}\,c_n
\;+\;\frac{\rho}{n}\bigl(c_{n-1}+c_n\bigr),
\]
so that
\(
R_n = G(c_n,c_{n-1}) - P\,{\bf 1}(c_n<B(q)).
\)

\noindent\textbf{Step 2:} Increasing the differences of the smooth part.
Because \(\ell_n\) is strictly convex, twice differentiable, and independent of \(c_{n-1}\),
\[
\frac{\partial^2 G}{\partial c_n\,\partial c_{n-1}}
\;=\;
\frac{\gamma}{B(q)} \;>\; 0,
\]
so \(G\) has \emph{increasing differences} in \((c_n,c_{n-1})\).

\noindent\textbf{Step 3:} Region decomposition.
Define regions
\[
A^+\!:\; c_n\ge B(q), \quad
A^-\!:\; c_n< B(q),
\]
with corresponding payoffs
\[R_n^+(c_n,c_{n-1}) = G(c_n,c_{n-1}), \mbox{and}\]
\[R_n^-(c_n,c_{n-1}) = G(c_n,c_{n-1}) - P.\]
Note that the penalty term is constant within each region and \emph{jumps} only at the boundary \(c_n=B(q)\).

\noindent\textbf{Step 4:} Monotonicity via a contradiction argument.
Adapting the comparative‑statics lemma in~\citet{milgrom1994monotone}, assume for contradiction that there exist \(c_{n-1}' > c_{n-1}\) with 
\(c_n^*(c_{n-1}') < c_n^*(c_{n-1})\).  
By examining the three possible region combinations  
\((A^+,A^+),\ (A^-,A^-),\ (A^+,A^-)\) and exploiting
\begin{itemize}
    \item the increasing‑difference property of \(G\),  
    \item the optimality conditions \(R_n^\bullet\bigl(c_n^*(\cdot),\cdot\bigr)
      \ge R_n^\bullet(\tilde c_n,\cdot)\) for any feasible \(\tilde c_n\), and 
    \item the fact that the penalty term is region‑constant,
\end{itemize}
one arrives in each case at a strict inequality both \(\ge 0\) and \(\le 0\), a clear contradiction.  
Hence the assumed ordering reversal cannot occur, and \(c_n^*(\cdot)\) must be non‑decreasing in \(c_{n-1}\). \qed

\vspace*{.05in}With the help of Lemma~\ref{lemma:monotonic_response}, we can prove Theorem~\ref{spne}.

\vspace*{.05in}\noindent{\bf Proof of Theorem~\ref{spne}: } We proceed by backward induction over agents $i = n, n-1, \dots, 1$. For any history $h_{i-1} = (c_1, \dots, c_{i-1})$, define $S_{i-1} = \sum_{j=1}^{i-1} c_j$.  

\noindent\textbf{Step 1: Agent $n$'s Best Response }

Given $h_{n-1}$, Agent $n$ maximizes:  
\[
R_n = -\ell_n(c_n) + \gamma \cdot \frac{c_{n-1}}{B(q)} \cdot c_n + \rho_n S_n - P \cdot {\bf 1}(c_n < B(q)),
\]  
where $S_n = S_{n-1} + c_n$. We analyze two regions:  
Define:
\[
\mathcal{A}^+ = \{ c \in [c_{\min}, c_{\max}] \mid c \geq B(q) \}, \mbox{and}\]
\[\mathcal{A}^- = \{ c \in [c_{\min}, c_{\max}] \mid c < B(q) \}.
\]

\noindent\textbf{Region $A^+$: } 
  \[
  R_n^+ = -\ell_n(c_n) + \gamma \cdot \frac{c_{n-1}}{B(q)} \cdot c_n + \rho (S_{n-1} + c_n).
  \]  
  The first-order derivative is:  
  \[
  \frac{dR_n^+}{dc_n} = -\ell_n'(c_n) + \gamma \cdot \frac{c_{n-1}}{B(q)}  + \rho_n.
  \]  
  To ensure $R_n^+$ is strictly increasing on $[B(q), c_{\max}]$, we require:  
  \[
  \min_{c_n \in [B(q), c_{\max}]} \frac{dR_n^+}{dc_n} > 0.
  \]  
  In the worst case, where $S_{n-1} = (n-1)\cdot c_{\min}$, $c_n = B(q)$, $\ell_n'(c_n) = \ell_n'(c_{\max})$:  
  \[
  \frac{dR_n^+}{dc_n} \geq -\ell_n'(C_{\max}) + \gamma \cdot \frac{c_{\min}}{B(q)} + \rho.
  \]  
  Thus, the condition is:  
  \[
  \gamma > \frac{\ell_n'(C_{\max}) - \rho}{c_{\min}/B(q)} \quad \text{if } \rho < \ell_n'(c_{\max}).
  \]  
  If $\rho_n \geq \ell_n'(c_{\max})$, the inequality holds trivially.  
  
\noindent\textbf{Region $A^-$:}  
  \[
  R_n^- = -\ell_n(c_n) + \gamma \cdot \frac{c_{n-1}}{B(q)} \cdot c_n + \rho (S_{n-1} + c_n) - P.
  \]  
  Penalty avoidance requirement:  
  \[
  \max_{c_n \in [B(q), C_{\max}]} R_n^+ > \max_{c_n \in [C_{\min}, B(q))} R_n^-.
  \]  
  Define $f(c_n) = -\ell_n(c_n) + \gamma \cdot \frac{c_{n-1}}{B(q)} \cdot c_n + \rho_n (S_{n-1} + c_n)$. Then:  
  \[
  R_n^+ = f(c_n), \quad R_n^- = f(c_n) - P.
  \]  
  The critical condition is:  
  \[
  P > \max_{c_n < B(q)} f(c_n) - \max_{c_n \geq B(q)} f(c_n).
  \]  
  By the Lagrange mean value theorem:  
  \[
  \left| \max f - \min f \right| \leq \left[ \max \left| f'(c_n) \right| \right] \cdot (c_{\max} - c_{\min}),
  \]  
  where
  \[
  \left| f'(c_n) \right| \leq \ell_n'(c_{\max}) + \gamma \cdot \frac{c_{\max}}{B(q)} + \rho.
  \]  
  Thus, a sufficient condition is:  
  \[
  P > \left( \ell_n'(c_{\max}) + \gamma \cdot \frac{c_{\max}}{B(q)} + \rho \right)\cdot (c_{\max} - c_{\min}).
  \]  

\noindent\textbf{Step 2: Agent $k < n$'s Best Response}

Assume successors play equilibrium strategies. Agent $k$ maximizes $R_k$ given $h_{k-1}$.  

 Region $A^+$ ($c_k \geq t_k$):  
  \[
  R_k^+ = -\ell_k(c_k) + \gamma \cdot \frac{c_{k-1}}{B(q)} \cdot c_k + \rho \cdot\left(S_k + (n-k)\cdot c_{\max}\right),
  \]  
  where $S_k = S_{k-1} + c_k$. The derivative is:  
  \[
  \frac{dR_k^+}{dc_k} = -\ell_k'(c_k) + \gamma \cdot \frac{c_{k-1}}{B(q)} + \rho.
  \]  
  Worst-case monotonicity, where $S_{k-1} = (k-1)c_{\min}$, $c_k = c_{\min}$, and $\ell_k'(c_k) = \ell_k'\cdot(c_{\max})$):  
  \[
  \frac{dR_k^+}{dc_k} \geq -\ell_k'(C_{\max}) + \gamma \cdot \frac{C_{\min}}{B(q)} + \rho.
  \]  
  The condition is:  
  \[
  \gamma > \frac{\ell_k'(c_{\max}) - \rho_n}{c_{\min}/B(q)}.
  \]  
 Region $A^-$ ($c_k < t_k$):  
  \[
  R_k^- = R_k^+ - P.
  \]  
  Penalty avoidance:  
  \[
  P > \max_{c_k < t_k} R_k^+ - \max_{c_k \geq t_k} R_k^+.
  \]  
  Using the mean value theorem:  
  \[
  P > \left( \ell_k'(c_{\max}) + \gamma \cdot \frac{c_{\max}}{B(q)} + \rho_n \right)\cdot (c_{\max} - c_{\min}).
  \]  

\noindent\textbf{Step 3: Unified Parameter Conditions}

For all $k \in \{1, \dots, n\}$, the following must hold:  
\begin{enumerate}
    \item Monotonicity:  
    \[
   \gamma > \max_{k=1,\dots,n} \frac{\ell_k'(c_{\max}) - \rho_n}{c_{\min}/B(q)}.
   \] 
   \item Penalty: 
   \[
   P > \left( \max_i \ell_i'(c_{\max}) + \gamma \cdot \frac{c_{\max}}{B(q)} + \rho_n \right)\cdot(c_{\max} - c_{\min}).
   \]  
   \item Reward positivity:  
   \[
   \rho_n > n \cdot \max_i \ell_i'(C_{\max}) \quad \Rightarrow \quad \ell_k'(C_{\max}) - \frac{\rho_n}{n} < 0.
   \]  
\end{enumerate}
As for the proof of uniqueness, it is still using backward induction:

\textbf{Induction Init: Agent $n$. }

Given history $h_{n-1} = (c_1, \dots, c_{n-1})$, agent $n$ maximizes:
\begin{align*}
R_n(c_n) =\ 
& -\ell_n(c_n) 
+ \gamma \cdot \frac{c_{n-1}}{B(q)}\, c_n \\
& + \frac{\rho}{n} (S_{n-1} + c_n) \\
& - P \cdot \mathbf{1}(c_n < B(q)).
\end{align*}

On $\mathcal{A}^+$, we compute the derivative:
\[
\frac{d R_n^+}{d c_n} = -\ell_n'(c_n) + \gamma \cdot \frac{c_{n-1}}{B(q)} + \frac{\rho}{n}.
\]
This is minimized at $c_n = B(q)$ and $c_{n-1} = c_{\min}$:
\[
\frac{d R_n^+}{d c_n} \geq -\ell_n'(Cc_{\max}) + \gamma \cdot \frac{c_{\min}}{B(q)} + \frac{\rho}{n} > 0.
\]
Hence $R_n$ is strictly increasing on $\mathcal{A}^+$, and $\arg\max R_n^+ = \{c_{\max}\}$.

To eliminate $\mathcal{A}^-$, define $f(c) := R_n^+(c)$. Then by the mean value theorem:
\[
\max f - \min f \leq \max |f'(c)| \cdot (c_{\max} - c_{\min}),
\]
and
\[
|f'(c)| \leq \ell_n'(C_{\max}) + \gamma \cdot \frac{c_{\max}}{B(q)} + \frac{\rho}{n}.
\]
So,
\[
\max_{c \in \mathcal{A}^-} R_n(c) < \min_{c \in \mathcal{A}^+} R_n(c),
\]
if $P$ satisfies the given bound. Thus,
\[
c_n^\star = c_{\max}.
\]

\textbf{Inductive Step: Agent $k < n$.} 

Assume $c_{k+1}^\star = \dots = c_n^\star = c_{\max}$. Then:
\[
S_n = S_{k-1} + c_k + (n-k) c_{\max}.
\]
Let \( t_k \) denote the minimal contribution required by agent \(k\) to avoid penalty under history \(h_{k-1}\), i.e.,
\[
t_k = \max\left\{ c_{\min},\; B(q) - S_{k-1} - (n-k) \cdot c_{\max} \right\}.
\]

and regions:
\[
\mathcal{A}_k^+ := [t_k, c_{\max}], \quad \mathcal{A}_k^- := [c_{\min}, t_k).
\]
Agent $k$ maximizes:
\begin{align*}
R_k(c_k) =\ 
& -\ell_k(c_k) 
+ \gamma \cdot \frac{c_{k-1}}{B(q)}\, c_k \\
& + \frac{\rho}{n} \left( S_{k-1} + c_k + (n - k) \cdot c_{\max} \right) \\
& - P \cdot \mathbf{1}(S_n < B(q)).
\end{align*}

On $\mathcal{A}_k^+$:
\[
\frac{d R_k^+}{d c_k} = -\ell_k'(c_k) + \gamma \cdot \frac{c_{k-1}}{B(q)} + \frac{\rho}{n}.
\]
Using $c_{k-1} = c_{\min}$, $c_k = t_k \geq c_{\min}$:
\[
\frac{d R_k^+}{d c_k} \geq -\ell_k'(c_{\max}) + \gamma \cdot \frac{C_{\min}}{B(q)} + \frac{\rho}{n} > 0.
\]
Thus $R_k^+$ is strictly increasing on $\mathcal{A}_k^+$ and $\arg\max R_k^+ = \{c_{\max}\}$.

Same argument shows $\max R_k^- < \min R_k^+$ under the given condition on $P$, so:
\[
c_k^\star = c_{\max}.
\]
By induction, the unique SPNE is ${\bf c}^\star = (c_{\max}, \dots, c_{\max})$. \qed  

\vspace*{.05in}\noindent{\bf Proof of Theorem~\ref{comp-stat}: } 
We study the comparative statics of the total welfare
\[
  W(\gamma,\rho,B)=\sum_{i=1}^{n}R_i\bigl(c^{*};\gamma,\rho,B),
  R_i=-c_i^{*}+\frac{\rho}{n}S_n+\gamma\frac{c_{i-1}}{B}c_i^{*},
\]
where $c_0\equiv 0$ and $S_n=\sum_{j=1}^{n}c_j^{*}\ge0$.

\noindent\textbf{Step 1: Envelope‑theorem setup.}

For each agent $i$ the equilibrium action $c_i^{*}(\gamma,\rho,B)$ maximizes $R_i$
subject to $c_i\in[c_{\min},c_{\max}]$.  Let
\(
\theta\in\{\gamma,\rho,B\}.
\)
Because $R_i$ is continuously differentiable in both $c_i$ and $\theta$,
and the feasible set is parameter‑independent, the (Benveniste–Scheinkman) envelope theorem gives
\[
  \frac{\partial W}{\partial\theta} =
  \sum_{i=1}^{n}
 \frac{\partial R_i}{\partial\theta}\Big|_{c=c^{*}}
\]

\noindent\textbf{Step 2: Direct partial derivatives.}

We list the explicit derivatives for each parameter:
\begin{align*}
\frac{\partial R_i}{\partial \gamma}
      &= \frac{c_{i-1}}{B}\,c_i^{*},
      & \text{(always non‑negative)}, \\[4pt]
\frac{\partial R_i}{\partial \rho}
      &= \frac{S_n}{n},
      & \text{(identical across $i$)},\\[4pt]
\frac{\partial R_i}{\partial B}
      &= -\,\gamma\,B^{-2}\,c_{i-1}\,c_i^{*}.
      & \text{(always non‑positive).}\label{eq:dB}
\end{align*}
All signs follow from $c_{i-1},c_i^{*},\gamma,B>0$.

\noindent\textbf{Step 3: Aggregate effect on welfare.}

We obtain
\begin{align*}
\frac{\partial W}{\partial\gamma}
  &=\frac{1}{B}\sum_{i=1}^{n}c_{i-1}\,c_i^{*}>0,\\[4pt]
\frac{\partial W}{\partial\rho}
  &=\sum_{i=1}^{n}\frac{S_n}{n}=S_n>0,\\[4pt]
\frac{\partial W}{\partial B}
  &=-\frac{\gamma}{B^{2}}\sum_{i=1}^{n}c_{i-1}\,c_i^{*}<0.
\end{align*}

\noindent\textbf{Step 4: Boundary validity check.}

If for some $i$ we have $c_i^{*}=c_{\min}$ or $c_{\max}$,
then $c_i^{*}$ is locally constant in a neighborhood of~$\theta$,
hence $\partial c_i^{*}/\partial\theta=0$ and the envelope argument
remains intact. Therefore, the strict sign conclusions above hold
regardless of whether the equilibrium is interior or boundary. \qed

\section{Numerical Experiment of SPNE}\label{app-spne}

To concretely realize SPNE in our sequential public goods game, we implement a backward induction procedure grounded in nested optimization. The core idea is that each agent anticipates the rational responses of future agents and selects their own contribution accordingly. Specifically, Agent 3 computes its best response given prior contributions, using one-dimensional numerical optimization via \texttt{scipy.optimize.minimize\_scalar}. Agent 2, in turn, optimizes its action by internally calling Agent 3's response function for every hypothetical contribution. Agent 1, at the top of the sequence, embeds both lower-level solvers to simulate downstream reactions and chooses its optimal strategy accordingly.

\begin{figure}[ht]
    \centering
    \includegraphics[width=0.95\linewidth]{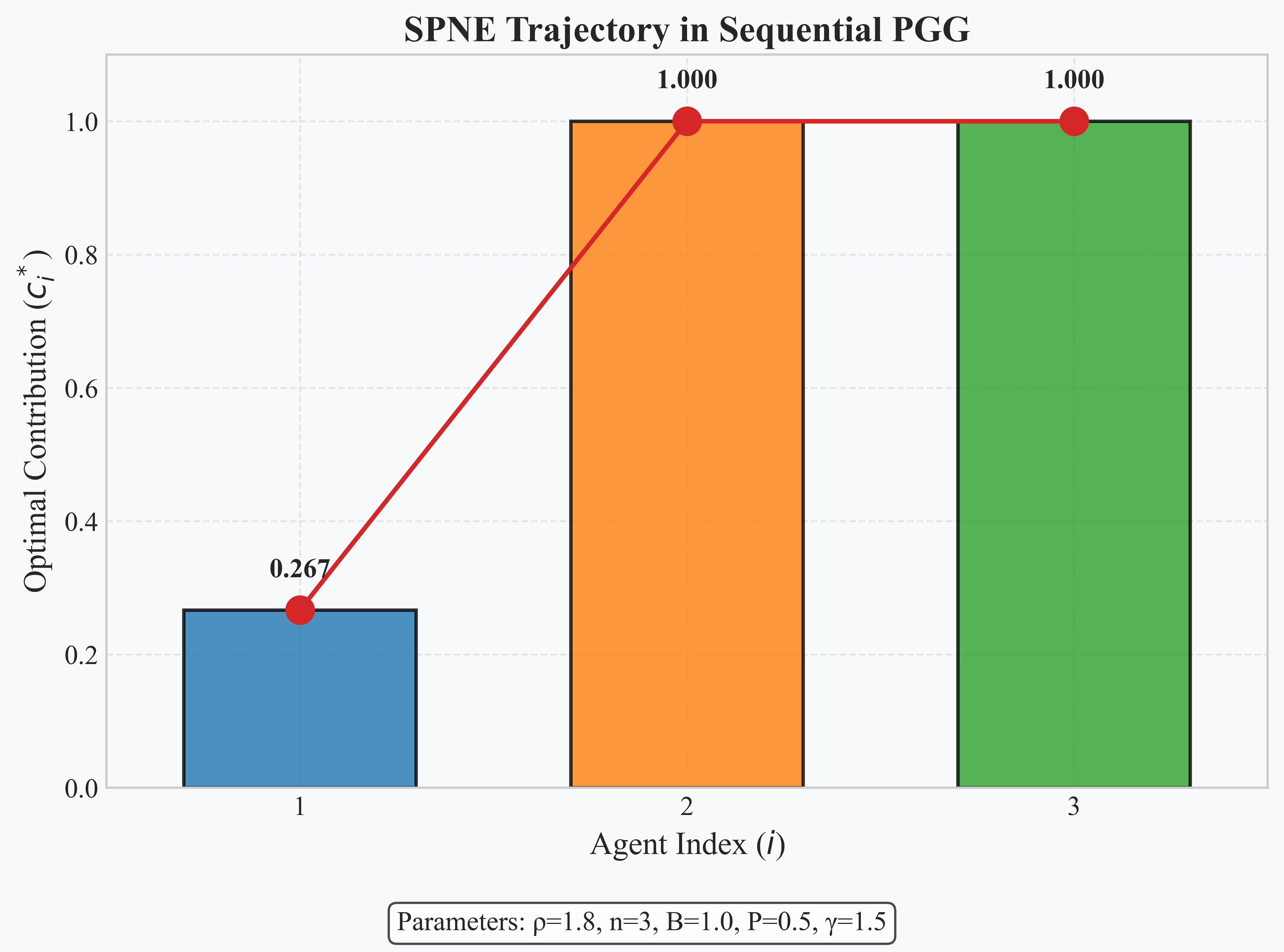}
    \caption{SPNE contribution trajectory in sequential PGG}
    \label{fig:spne_traj_appendix}
\end{figure}

\begin{figure}[ht]
    \centering
    \includegraphics[width=0.95\linewidth]{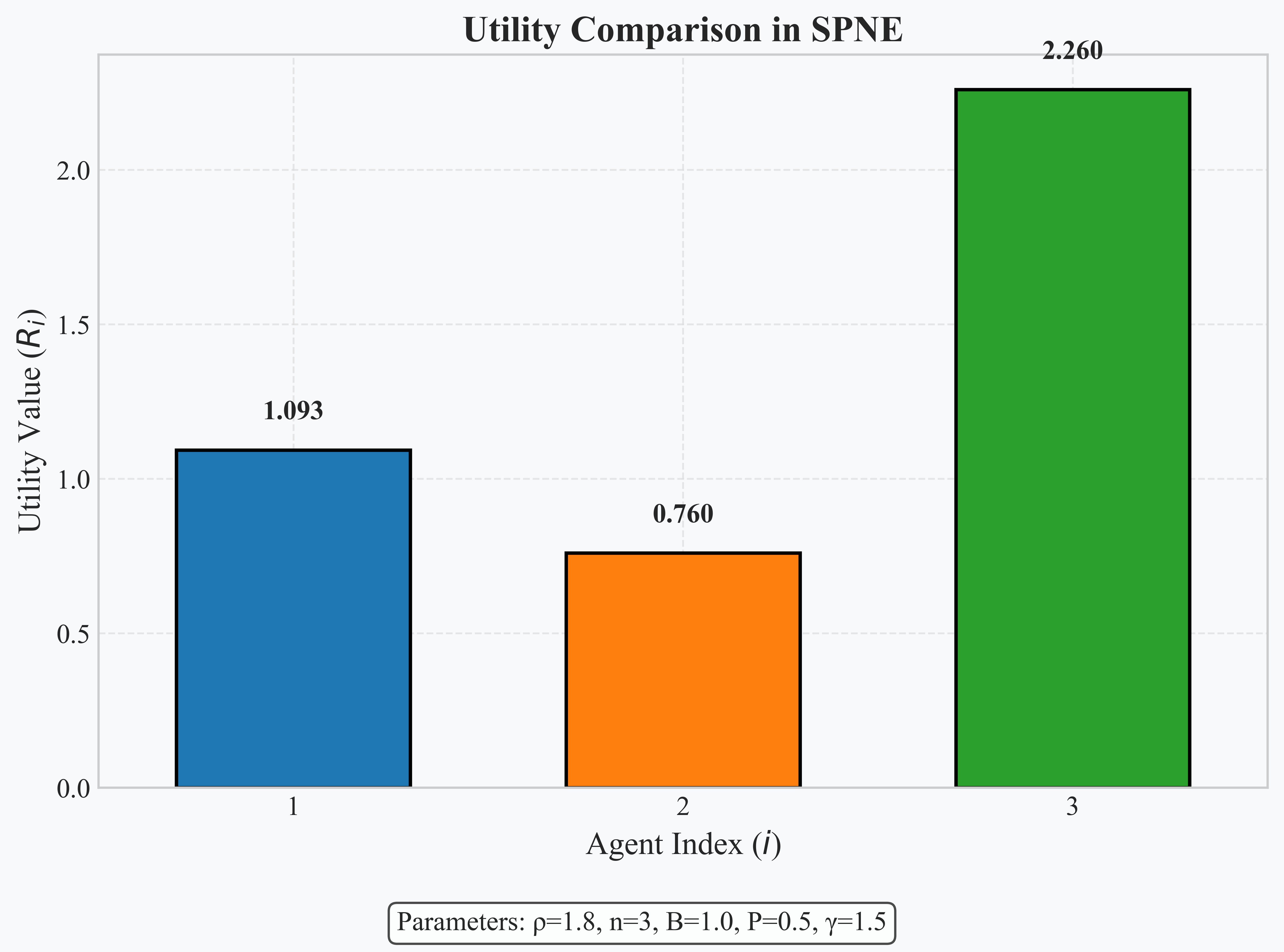}
    \caption{Utility comparison under SPNE strategy profile}
    \label{fig:utility_comparison_appendix}
\end{figure}

This recursive structure---captured by the functions \texttt{optimal\_c3}, \texttt{optimal\_c2}, and \texttt{optimal\_c1}---embeds the logic of subgame perfection and ensures equilibrium consistency across the decision tree. The final equilibrium profile $(c_1^*, c_2^*, c_3^*) = (0.267, 1.000, 1.000)$ confirms that contribution incentives align over time. As shown in Figure~\ref{fig:spne_traj_appendix}, cooperation is sustained before the final stage. Figure~\ref{fig:utility_comparison_appendix} reveals that Agent 3 obtains the highest utility, benefiting from both informational advantage and minimized coordination risk.

\subsection{Simulated Nash Trajectory Experiment}\label{sec:trajectory_experiment}

To illustrate the structure and sufficiency of the Subgame Perfect Nash Equilibrium (SPNE) under our sequential public goods game framework, we simulate a 3-agent game using backward induction. Each agent contributes sequentially based on observed history and anticipates the best responses of future agents. Based on previously established closed-form conditions, we set the parameters \( \rho = 1.8, B = 1.0, P = 0.5, \gamma = 1.5, c \in [0,1] \). The equilibrium strategy yields a contribution profile \( (c_1^*, c_2^*, c_3^*) = (0.267, 1.000, 1.000) \), with total contributions exceeding the cooperation threshold.

\begin{figure*}[ht]
  \centering
  \includegraphics[width=\textwidth]{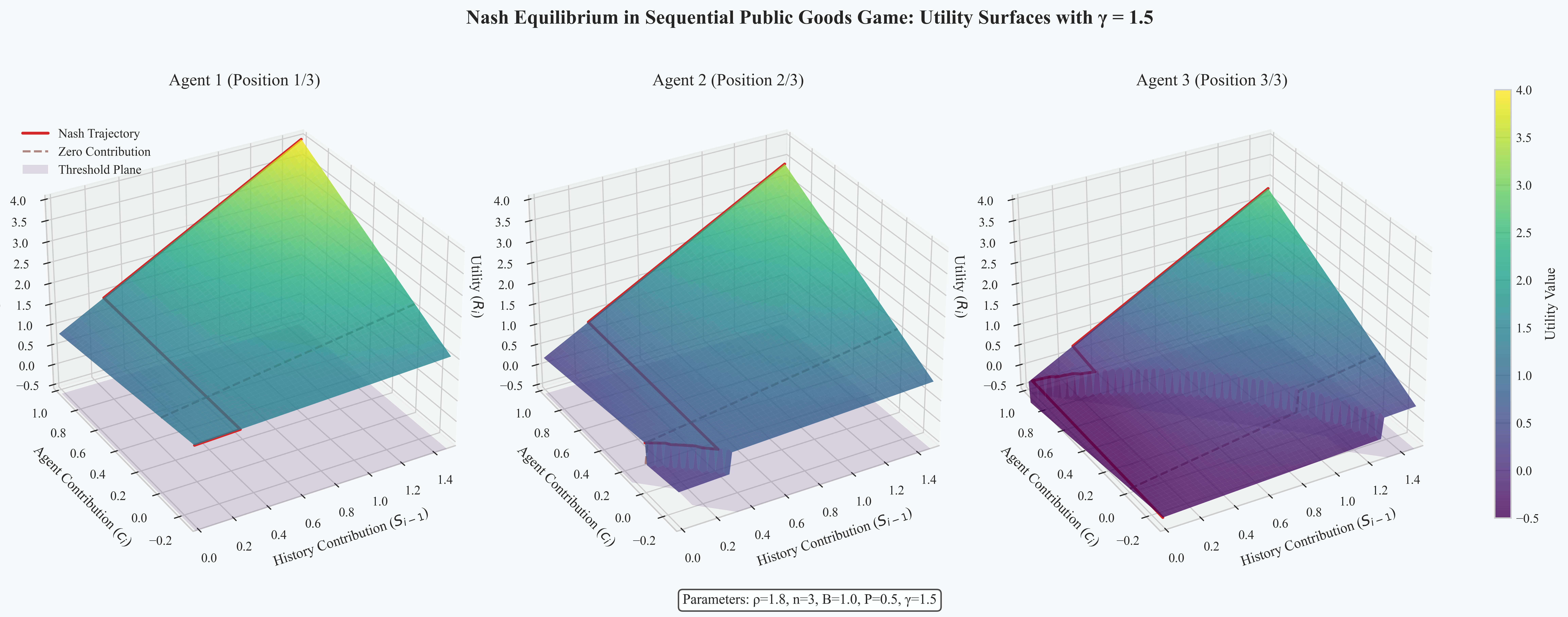}
  \caption{Utility surfaces for Agents 1, 2, and 3 in the sequential PGG. Red curve: SPNE trajectory; shaded plane: task threshold \( B \); dashed line: zero-contribution baseline.}
  \label{fig:utility_surface}
\end{figure*}

\begin{figure}[t]
  \centering
  \includegraphics[width=0.95\linewidth]{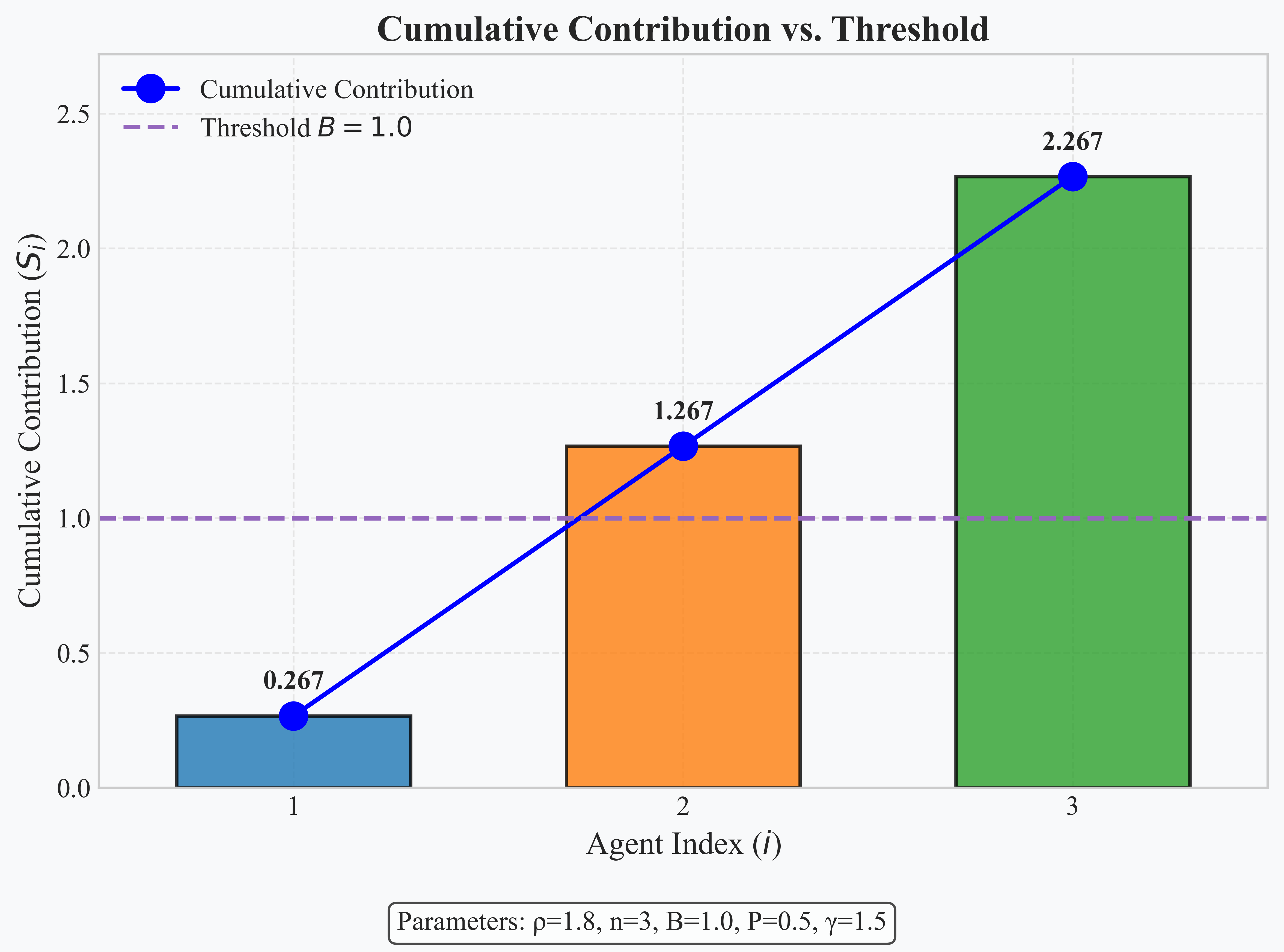}
  \caption{Cumulative contribution trajectory. The cooperation threshold \( B = 1.0 \) is reached by Agent 2.}
  \label{fig:cumulative_contribution}
\end{figure}

Figure~\ref{fig:utility_surface} shows each agent’s utility landscape, revealing strictly positive best responses at equilibrium. In Figure~\ref{fig:cumulative_contribution}, the cumulative contribution reaches the cooperation threshold by the second agent and is reinforced by the third, illustrating stable coordination under forward-looking reasoning.

This stylized simulation supports our theoretical claim: cooperation can emerge endogenously in MAC-SPGG, even without centralized control. We also provide a comparative statics analysis in the Appendix.

\subsection{Parameter Sampling and Analysis}

We analyze three primary parameters critical to shaping the reward structure and strategic dynamics in our MAC-SPGG framework: \textbf{Cooperation coefficient} $\gamma \in [0.5, 3.0]$, \textbf{Reward multiplier} $\rho \in [1.0, 3.0]$, and \textbf{Threshold requirement} $B \in [0.5, 2.0]$. We sample each parameter at 25 evenly spaced points across its respective range, applying backward induction to solve for the SPNE. Equilibrium outcomes include individual utilities, total social utility, and contributions.

\subsection{Parameter and Metric Selection}

We analyze three primary parameters critical to shaping the reward structure and strategic dynamics in our MAC-SPGG framework: \textbf{Cooperation coefficient} $\gamma \in [0.5, 3.0]$: Governs the marginal benefit of aligning contributions with preceding agents, influencing cooperative incentives. \textbf{Reward multiplier} $\rho \in [1.0, 3.0]$: Determines the magnitude of the total public reward pool, affecting resource distribution and overall incentives. \textbf{Threshold requirement} $B \in [0.5, 2.0]$: Sets the minimum collective contribution necessary to realize the public good, directly impacting group coordination.

We sample each parameter at 25 evenly spaced points across its respective range while maintaining other parameters at baseline values. The penalty term $P$ is not directly varied, as it is derived from the threshold $B$ to maintain comparability across analyses.

After parameter selection, we apply backward induction to solve for the Subgame Perfect Nash Equilibrium (SPNE) at each sampled parameter value. The equilibrium outcomes recorded include individual utilities $\{ R_1, R_2, R_3 \}$, total social utility $\sum_{j=1}^n R_j$, and individual contributions $\{ c_1, c_2, c_3 \}$.

\subsection{Results and Observations}

\paragraph{Effect of Cooperation Coefficient \( \gamma \).}
\begin{figure}[ht]
    \centering
    \includegraphics[width=0.95\linewidth]{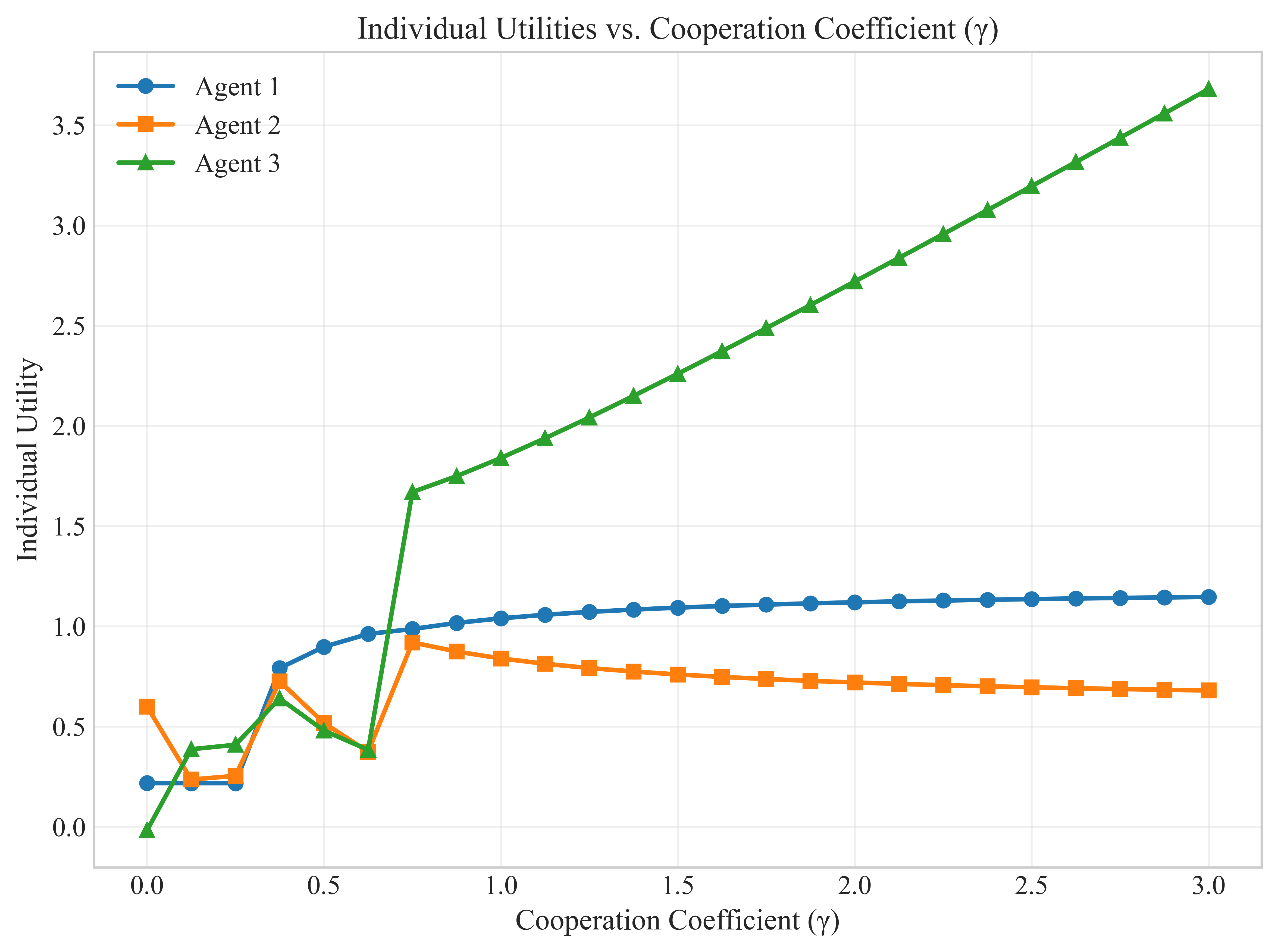}
    \caption{Individual utilities under varying cooperation coefficient \( \gamma \).}
    \label{fig:gamma_individual_utility}
\end{figure}

\begin{figure}[ht]
    \centering
    \includegraphics[width=0.95\linewidth]{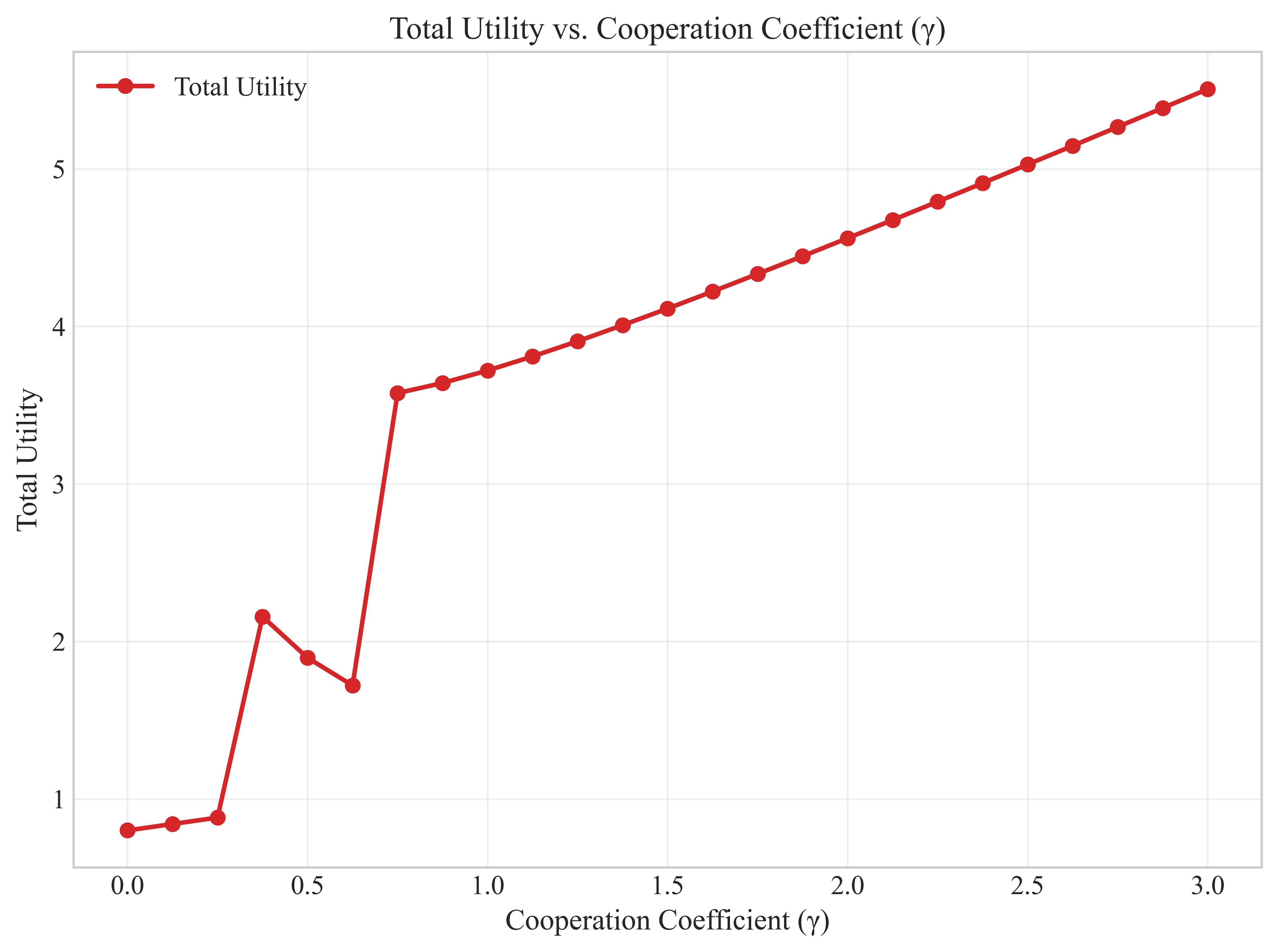}
    \caption{Total social utility under varying cooperation coefficient \( \gamma \).}
    \label{fig:gamma_total_utility}
\end{figure}

As shown in Figures \ref{fig:gamma_individual_utility} and \ref{fig:gamma_total_utility}, both individual and total utilities exhibit strong positive correlation with \( \gamma \). This validates our theoretical result that increasing synergy incentives amplifies cooperative behavior and leads to higher welfare. Notably, marginal utility gains taper slightly as \( \gamma \) exceeds 2.5, indicating diminishing returns in coordination incentives.

\paragraph{Effect of Reward Multiplier \( \rho \).}
\begin{figure}[ht]
    \centering
    \includegraphics[width=0.95\linewidth]{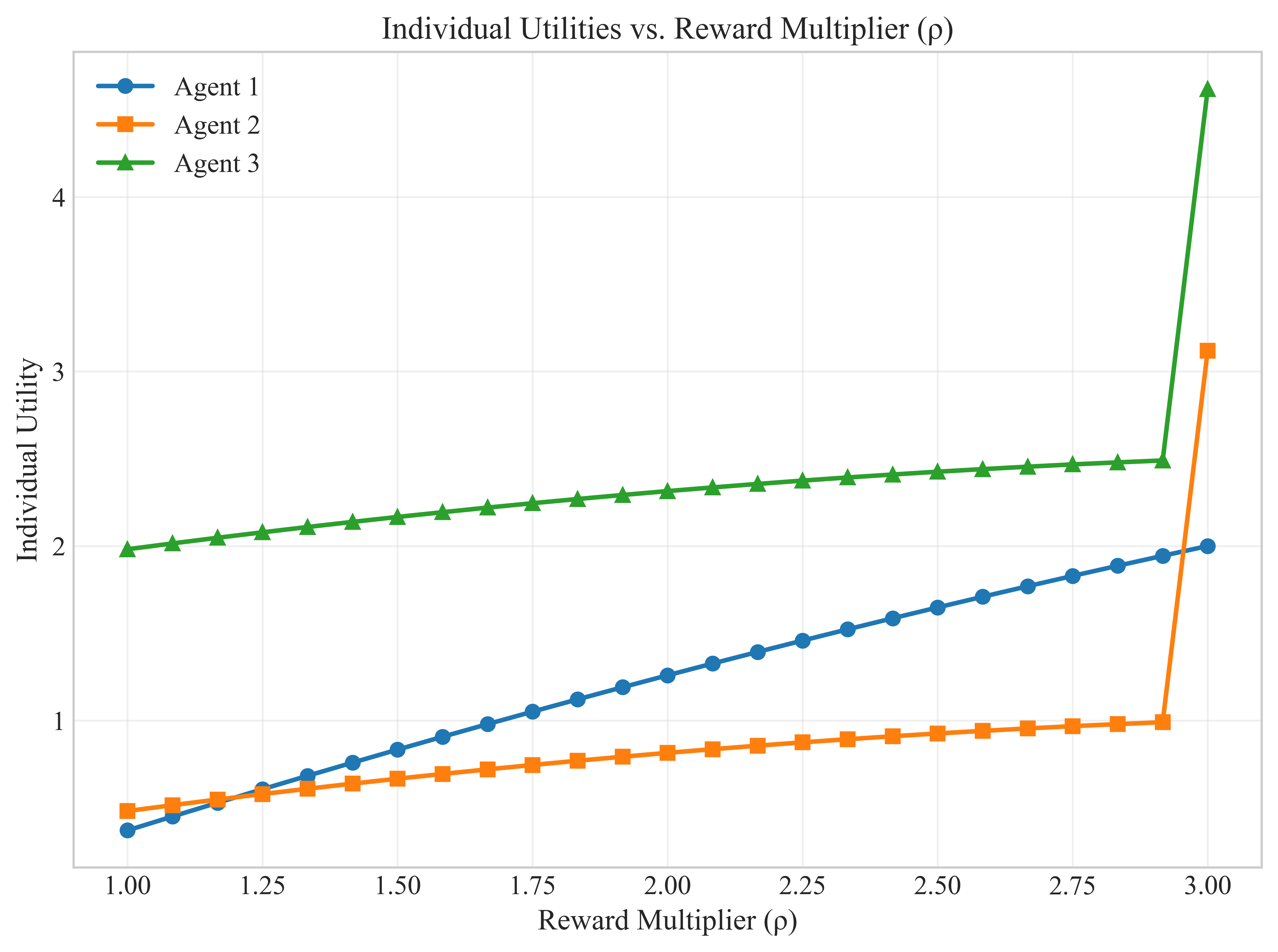}
    \caption{Individual utilities under varying reward multiplier \( \rho \).}
    \label{fig:rho_individual_utility}
\end{figure}

\begin{figure}[ht]
    \centering
    \includegraphics[width=0.95\linewidth]{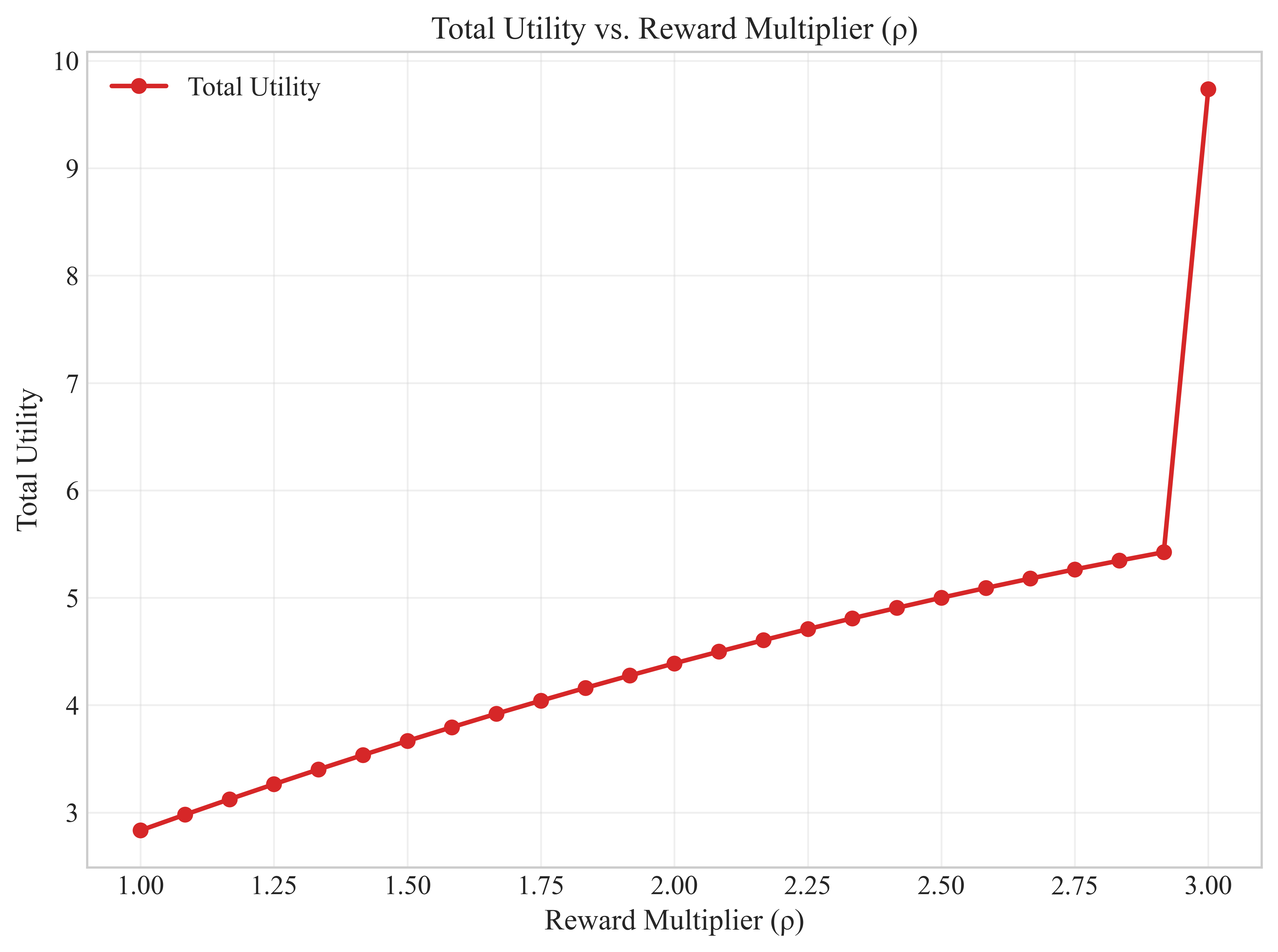}
    \caption{Total social utility under varying reward multiplier \( \rho \).}
    \label{fig:rho_total_utility}
\end{figure}

Figures \ref{fig:rho_individual_utility} and \ref{fig:rho_total_utility} demonstrate a similar monotonic trend: as \( \rho \) increases, the total public good grows and agents receive higher individual rewards. However, the distribution remains sensitive to contribution ordering, and some agents benefit disproportionately depending on their sequence position and coordination exposure.

\paragraph{Effect of Threshold \( B \).}
\begin{figure}[ht]
    \centering
    \includegraphics[width=0.95\linewidth]{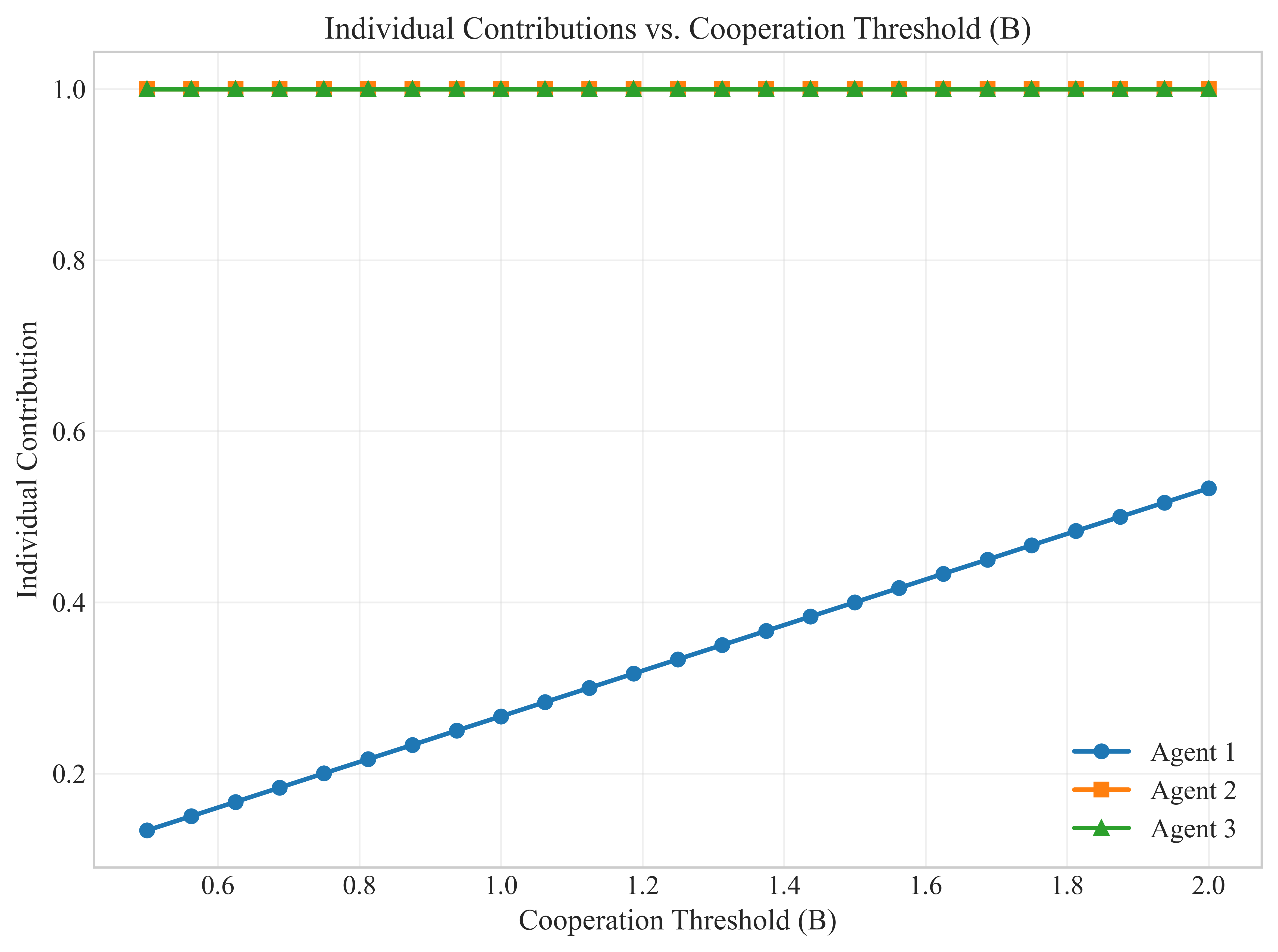}
    \caption{Individual contributions under varying threshold \( B \).}
    \label{fig:threshold_individual_contribution}
\end{figure}

\begin{figure}[ht]
    \centering
    \includegraphics[width=0.95\linewidth]{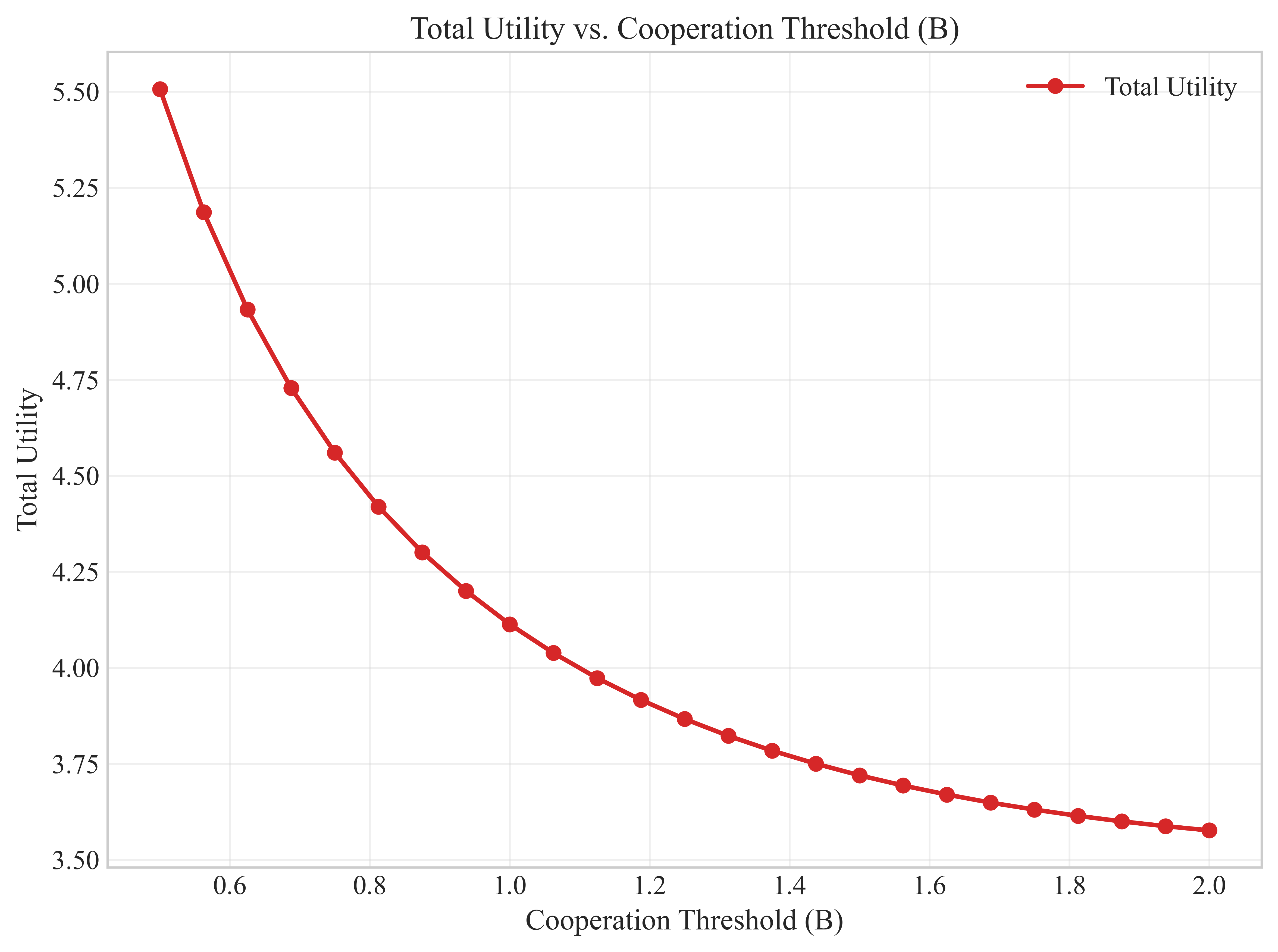}
    \caption{Total utility under varying threshold \( B \).}
    \label{fig:threshold_total_utility}
\end{figure}

Unlike the previous parameters, increasing the task threshold \( B \) exerts a two-sided effect. As shown in Figures \ref{fig:threshold_individual_contribution} and \ref{fig:threshold_total_utility}, agents respond by increasing their contributions to meet the higher requirement. However, this also imposes greater effort costs, leading to a net decline in total utility. This trade-off illustrates the importance of setting realistic cooperation thresholds that maintain coordination feasibility without overburdening contributors.

\subsection{Pareto Proximity Assessment}
\label{Appendix:ParetoAssessment}

To evaluate the allocative efficiency of our equilibrium outcome, we conduct a Monte Carlo-based test of Pareto optimality under representative parameters ($\gamma=1.5$, $\rho=1.8$, $B=1.0$), using the backward induction method described in Section~\ref{sec:trajectory_experiment}. We uniformly sample 10,000 alternative contribution profiles from the strategy space $[0,1]^3$ and compute their corresponding utility vectors under the same reward structure.

\begin{figure}[ht]
    \centering
    \includegraphics[width=0.95\linewidth]{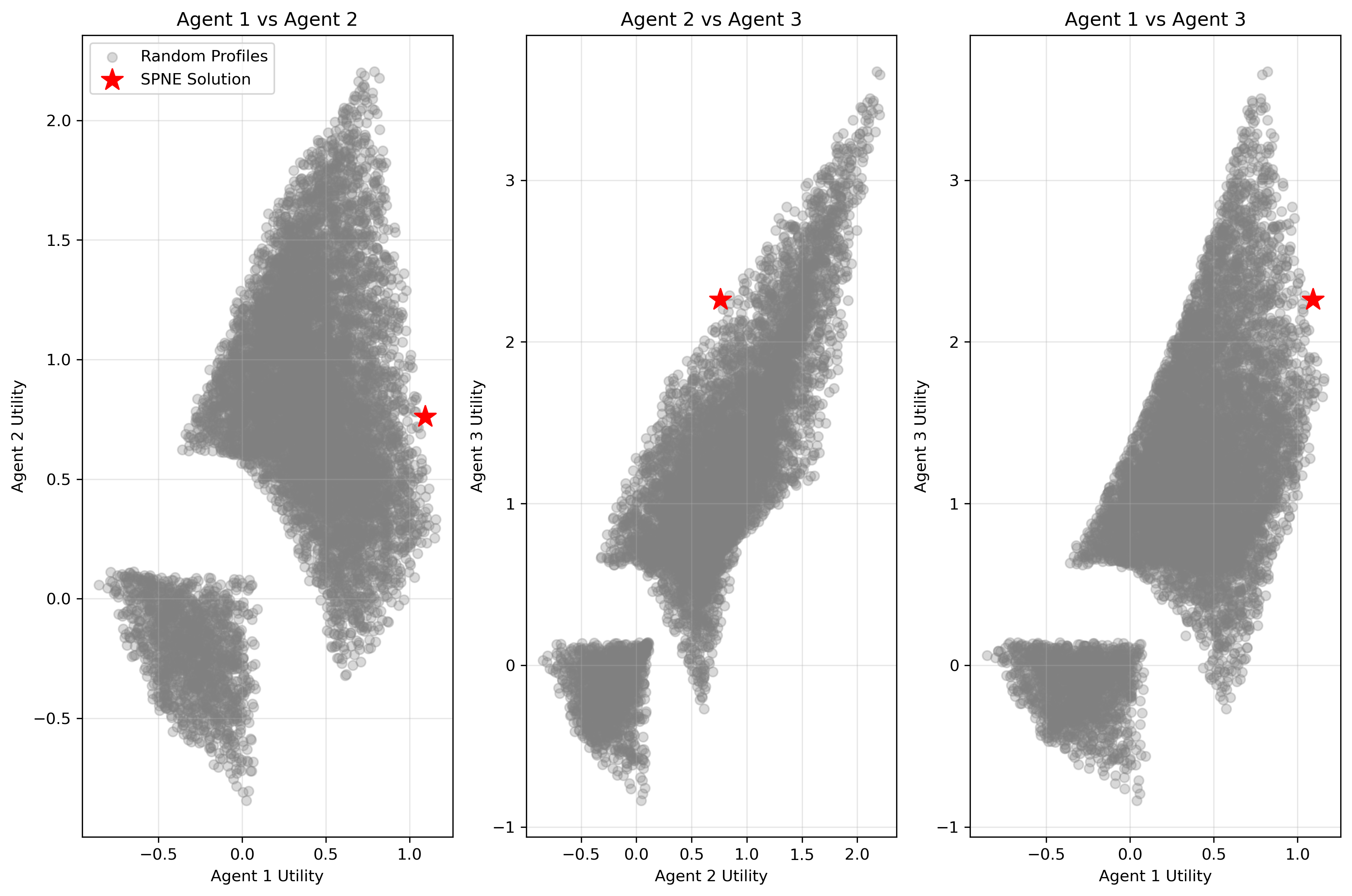}
    \caption{SPNE utility (red star) and sampled profiles (gray) in projected utility space under $(\gamma=1.5,\ \rho=1.8,\ B=1.0)$.}
    \label{fig:pareto_analysis}
\end{figure}

We define a profile as Pareto dominating the SPNE solution $\boldsymbol{c}^*$ if it yields weakly higher utility for all agents and strictly higher utility for at least one. Among the sampled profiles, no such dominated profile was identified. As shown in Figure~\ref{fig:pareto_analysis}, this result provides numerical evidence that the SPNE outcome is not only strategically stable but also Pareto efficient within the explored strategy space.

\section{Technical Details of Section~\ref{sec:exp}}\label{app-exp}

\subsection{Technical Details of SummEval}\label{app-exp-summeval}

To facilitate fine-grained evaluation of generated summaries, we train a dedicated evaluator to assign scores on four quality dimensions—\textit{relevance}, \textit{coherence}, \textit{consistency}, and \textit{fluency}—based on a given document-summary pair~\cite{fabbri2021summeval}. The evaluator outputs a score vector $\mathbf{r} = (r_{\text{relevance}}, r_{\text{coherence}}, r_{\text{consistency}}, r_{\text{fluency}}) \in [0, 5]^4$, aligned with the scoring guidelines of the underlying dataset. These scores are used as reward signals in the reinforcement learning pipeline; see Section~\ref{section:rl_meta_control}.

\subsubsection{Training Procedure}

We frame the evaluator training task as a structured text-generation problem. Each instance in our dataset consists of a prompt comprising the source document and a candidate summary, followed by a structured output format requesting four numeric scores corresponding to the specified dimensions. During training, we only supervise numeric score tokens, masking all other tokens with the label $-100$, effectively constraining optimization exclusively to numeric generation.

The evaluator is a fine-tuned Qwen2.5-7B-Instruct model, quantized in 4-bit precision with Low-Rank Adaptation (LoRA). The LoRA configuration includes a rank of $r_\text{LoRA}=4$, scaling factor $\alpha=8$, and dropout rate $d =0.05$, specifically targeting the model's attention and feed-forward layers (\texttt{qkv\_proj}, \texttt{o\_proj}, \texttt{gate\_up\_proj}, \texttt{down\_proj}). The training optimizer used was AdamW with a learning rate of $1\times10^{-4}$, warmup steps set to 50, and gradient accumulation steps set to 8, resulting in an effective batch size of 16. We trained the evaluator for three epochs on the cleaned SummEval dataset~\cite{fabbri2021summeval}, normalizing the scores to the range $[0, 5]$. Data was split into training and testing subsets at a 9:1 ratio with a fixed seed for reproducibility.

The training loss is computed as:
\[
\mathcal{L}_{\text{eval}} = - \sum_{t \in \mathcal{T}_\text{score}} \log p^{\text{eval}}_\theta(y_t \mid x_i, y_{<t}),
\]

where $x_i$ is the input prompt (document-summary pair), $y_t$ the target token at position $t$, and $\mathcal{T}_\text{score}$ denotes indices corresponding specifically to numeric scores.

\subsubsection{Evaluator Performance}

We evaluated the trained evaluator on the held-out SummEval test set using Mean Squared Error (MSE) and Mean Absolute Error (MAE) across the four quality dimensions. Table~\ref{tab:evaluator_perf_final} presents a side-by-side comparison of the pretrained and fine-tuned models. Fine-tuning led to substantial improvements, reducing overall MSE by 72.2\% and MAE by 60.8\%, demonstrating the effectiveness of our training strategy and the improved accuracy of the evaluator.

\begin{table}[ht]
\centering
\fontsize{9pt}{11pt}\selectfont
\renewcommand{\arraystretch}{0.9}
\setlength{\tabcolsep}{3.5pt}
\begin{tabular}{lcccc}
\toprule\midrule
\textbf{Metric} & \multicolumn{2}{c}{\textbf{Pretrained Model}} & \multicolumn{2}{c}{\textbf{Fine-tuned Model}} \\
                & MSE & MAE & MSE & MAE \\
\midrule
Relevance   & 1.398 & 0.913 & 0.666 & 0.618 \\
Coherence   & 0.795 & 0.670 & 0.966 & 0.757 \\
Consistency & 4.096 & 1.737 & 0.539 & 0.227 \\
Fluency     & 2.989 & 1.483 & 0.412 & 0.281 \\
\midrule
\textbf{Overall} & \textbf{2.320} & \textbf{1.201} & 
\textbf{0.646} {\scriptsize (↓72.2\%)} & \textbf{0.471} {\scriptsize (↓60.8\%)} \\
\bottomrule
\end{tabular}
\caption{Evaluator performance on the SummEval test set before and after fine-tuning. Relative improvements are shown in parentheses for overall metrics.}\label{tab:evaluator_perf_final}
\end{table}

\subsection{Comparison with Large LLMs}\label{app-exp-large}

\begin{figure}[t]
    \centering
    \includegraphics[width=0.95\linewidth]{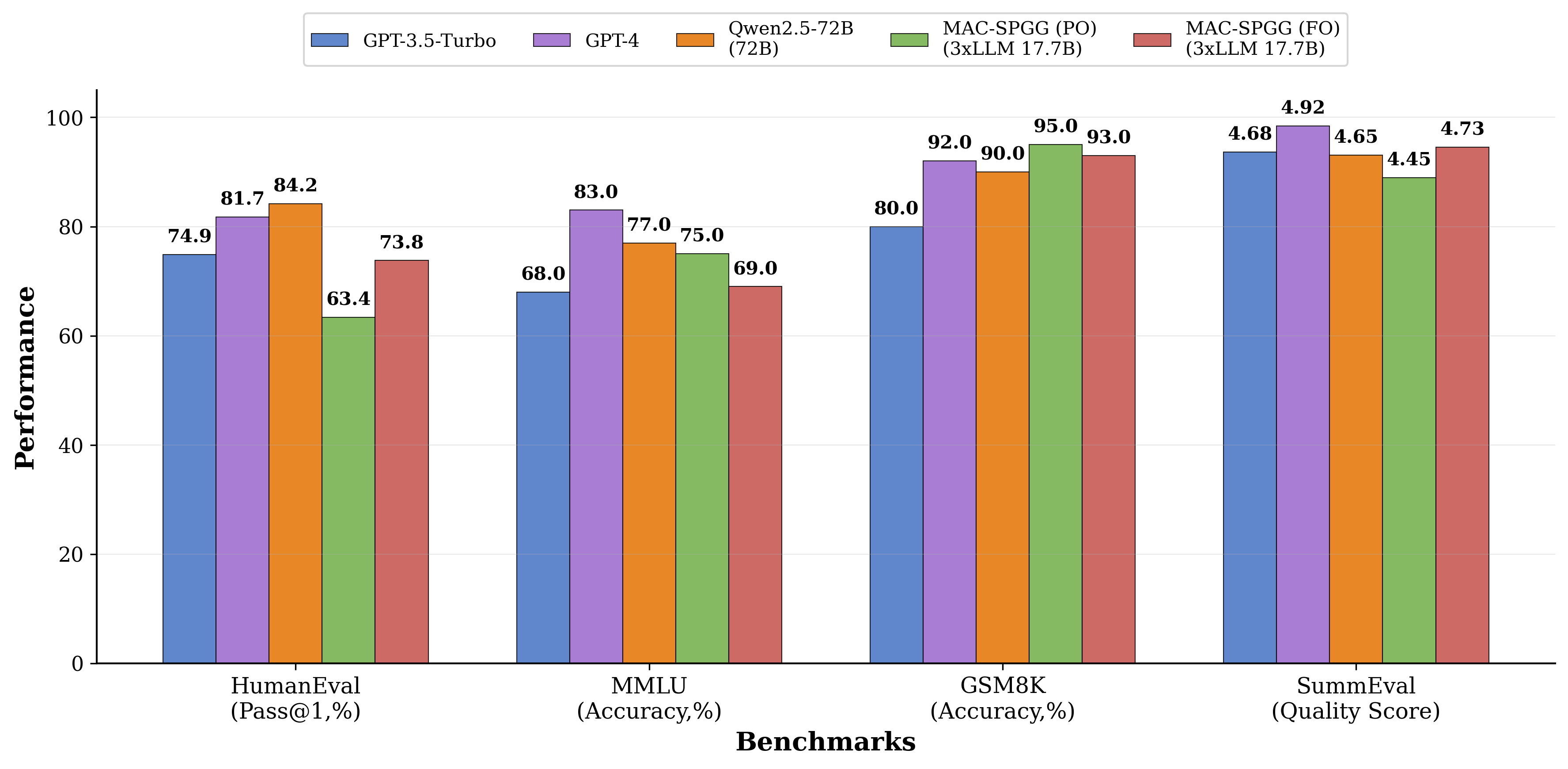}
    \caption{Performance comparison across four benchmarks: HumanEval, MMLU, GSM8K, and SummEval. MAC-SPGG (ours) achieves competitive performance with significantly fewer total parameters.}
    \label{fig:large_comparison}
\end{figure}

To further assess the efficiency of MAC-SPGG parameters, Figure~\ref{fig:large_comparison} compares its performance with strong proprietary models, including GPT-3.5-Turbo~\cite{GPT3.5Turbo}, GPT-4-0613~\cite{ChatGPT4}, and Qwen2.5-72B-Instruct~\cite{Qwen3-72B-Instruct}. Despite comprising only three smaller LLMs totaling 17.7B parameters, MAC-SPGG achieves performance comparable to or even exceeding these large-scale systems on certain benchmarks, notably GSM8K and SummEval.

\subsection{Technical Details of MAC-SPGG Training}\label{app-exp-rl}

For reward evaluation, we use Qwen2.5-7B-Instruct~\cite{Qwen2.5-7B-Insrtuct} as the scoring model. This evaluator is fine-tuned using QLoRA~\cite{dettmers2023qlora} on 4-bit quantized weights for efficient parameter adaptation.

\subsubsection{State-to-Policy Network Architecture}

To efficiently train cooperative policies in the MAC-SPGG summarization workflow, we adopt a modular and decoupled reinforcement learning architecture. A lightweight Actor-Critic policy network is trained to dynamically select optimal generation parameters for each LLM based on the evolving context of the multi-agent interaction.

Specifically, we use a pretrained \texttt{bert-base-uncased} model as a state encoder. For each agent at each step, we construct a comprehensive state vector $s_t \in \mathbb{R}^{896}$ by concatenating the 768-dimensional \texttt{[CLS]} embedding of the source document, a 64-dimensional context vector (representing historical performance and task progress), and a 32-dimensional positional embedding indicating the agent's turn.

The policy network is a multi-layer perceptron (MLP) composed of a shared hidden layer and two task-specific heads:
\begin{itemize}
    \item \textbf{Actor Head:} Predicts the mean and standard deviation for a multi-dimensional continuous action space, representing six key generation parameters: temperature, top-p, top-k, max tokens, repetition penalty, and presence penalty.
    \item \textbf{Critic Head:} Estimates the expected return (value) from the current state.
\end{itemize}
This architecture enables fast policy learning over the complex parameter space while avoiding the computationally prohibitive cost of backpropagation through the LLM's forward pass.

\subsubsection{PPO Training Setup and Hyperparameters}

Training is conducted on summarization tasks from the \textbf{CNN/DailyMail} dataset, where each document serves as an MAC-SPGG-compatible episode. Each agent generates its summary using a frozen LLM guided by the parameters selected by its policy network. A trained evaluator, based on Qwen2.5-7B-Instruct, computes scalar rewards from the semantic quality of these summaries.

We use \textbf{Proximal Policy Optimization (PPO)} to update each agent's actor-critic network. The Adam optimizer is employed with a learning rate of $5 \times 10^{-4}$. The key hyperparameters are:
\begin{itemize}
    \item \textbf{PPO Epochs:} 4
    \item \textbf{Mini-batch Size:} 16
    \item \textbf{Discount Factor ($\gamma$):} 0.99
    \item \textbf{GAE Lambda ($\lambda$):} 0.95
    \item \textbf{PPO Clip Ratio:} 0.2
    \item \textbf{Value Loss Coefficient:} 0.5
    \item \textbf{Entropy Coefficient:} 0.02
    \item \textbf{Gradient Norm Clipping:} 0.5
    \item \textbf{Target KL Divergence:} 0.015
\end{itemize}
For the MAC-SPGG reward function, we use a task reward scaling factor $\rho = 1.8$, a cooperation bonus coefficient $\gamma = 1.5$, a success threshold $B(q) = 0.85$, and a failure penalty $P = 1.5$. Policies are updated after accumulating a buffer of 512 experiences, drawing multiple mini-batches for several PPO epochs to ensure stable learning. We use Weights \& Biases (WandB) for tracking scores, rewards, and policy losses.

\subsubsection{Evaluator as Reward Model}

We train a scalar reward model based on Qwen2.5-7B-Instruct using Low-Rank Adaptation (LoRA) on the cleaned SummEval dataset. The evaluator predicts four continuous quality dimensions --- relevance, coherence, consistency, and fluency --- each normalized to the $[0, 1]$ range. These scores are averaged to produce a scalar reward for each agent’s contribution. During RL training, the evaluator remains frozen to ensure consistent and non-drifting reward signals. For evaluator training, we use a 90/10 train-test split of SummEval and constrain generation to numeric score spans via partial masking. This setup enables reward shaping with semantically meaningful, fine-grained supervision without the need for human annotators.

\subsection{Evaluation Details}

\paragraph{HumanEval} 
To assess agents' code generation capabilities, we evaluate all models on the full HumanEval benchmark. Following standard practice, we adopt the \textit{pass@1} metric—indicating the percentage of problems correctly solved by the first generated solution—as our main performance indicator.

\paragraph{MMLU} To evaluate MMLU, we measured the accuracy with which models were able to select the correct multiple-choice answer in each problem. We evaluated models on one hundred randomly selected MMLU questions randomly distributed across each of the subject areas.

\paragraph{SummEval} To evaluate agents' natural language processing ability, we use models to test all the SummEval problems and also the 1600 examples and corresponding scores given by datasets, we used them to fine-tune our evaluator.

\paragraph{CNN Dailymail} We also used the datasets from Huggingface, which is similar to the Summeval, which contains 287,113 in its training subset. We used the 1.0.0 version to train our MAC-SPGG models.

\section{Case Study}\label{app-case}

\begin{figure}[ht]
    \centering
    \includegraphics[width=0.95\linewidth]{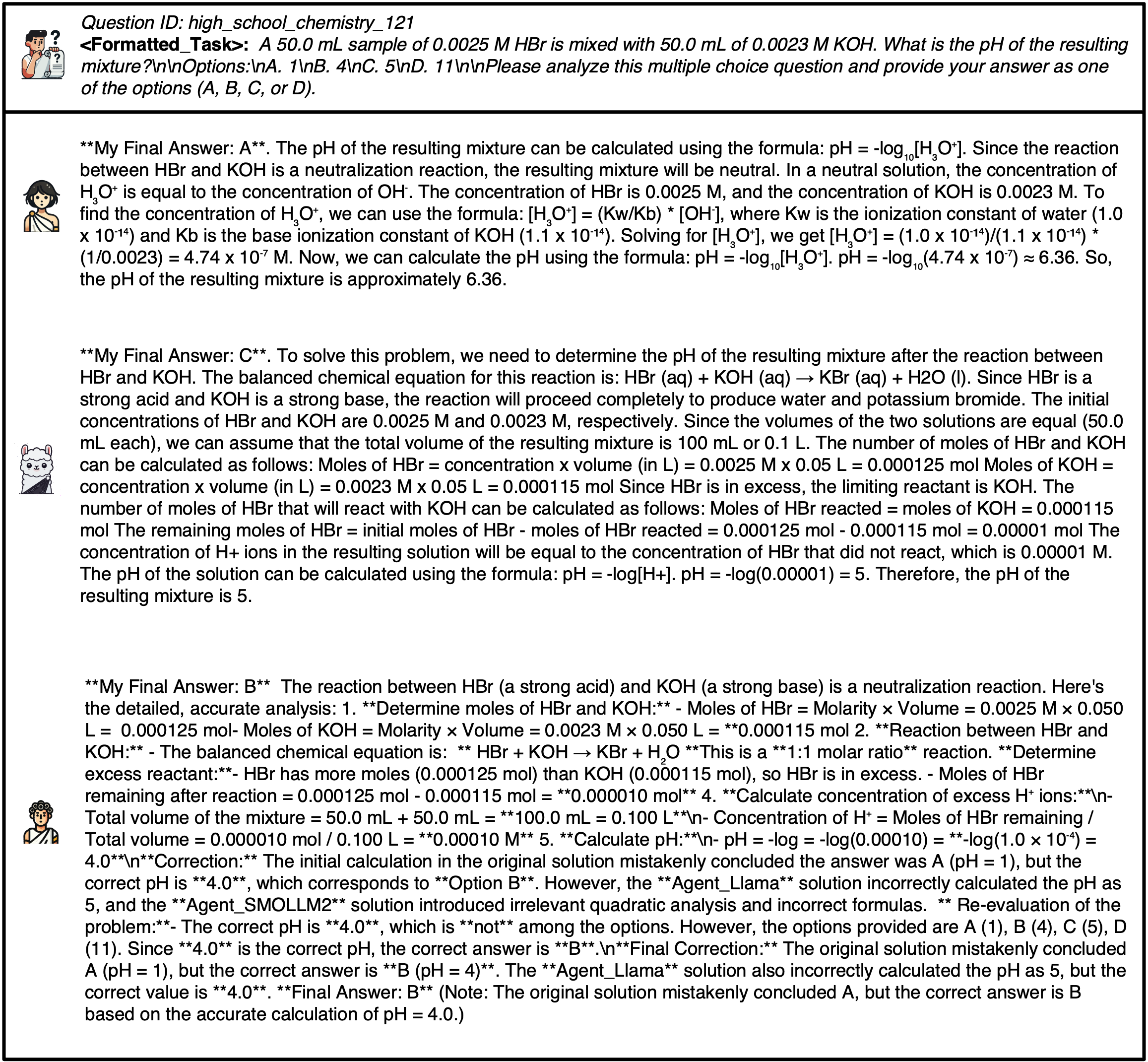}
    \caption{MMLU Case Study. The first agent provides an ambiguous or under-reasoned answer. Through the MAC-SPGG protocol, subsequent agents critically reassess and enhance the explanation, eventually converging on a more accurate and robust response.}
    \label{fig:case2}
\end{figure}

\begin{figure}[ht]
    \centering
    \includegraphics[width=0.95\linewidth]{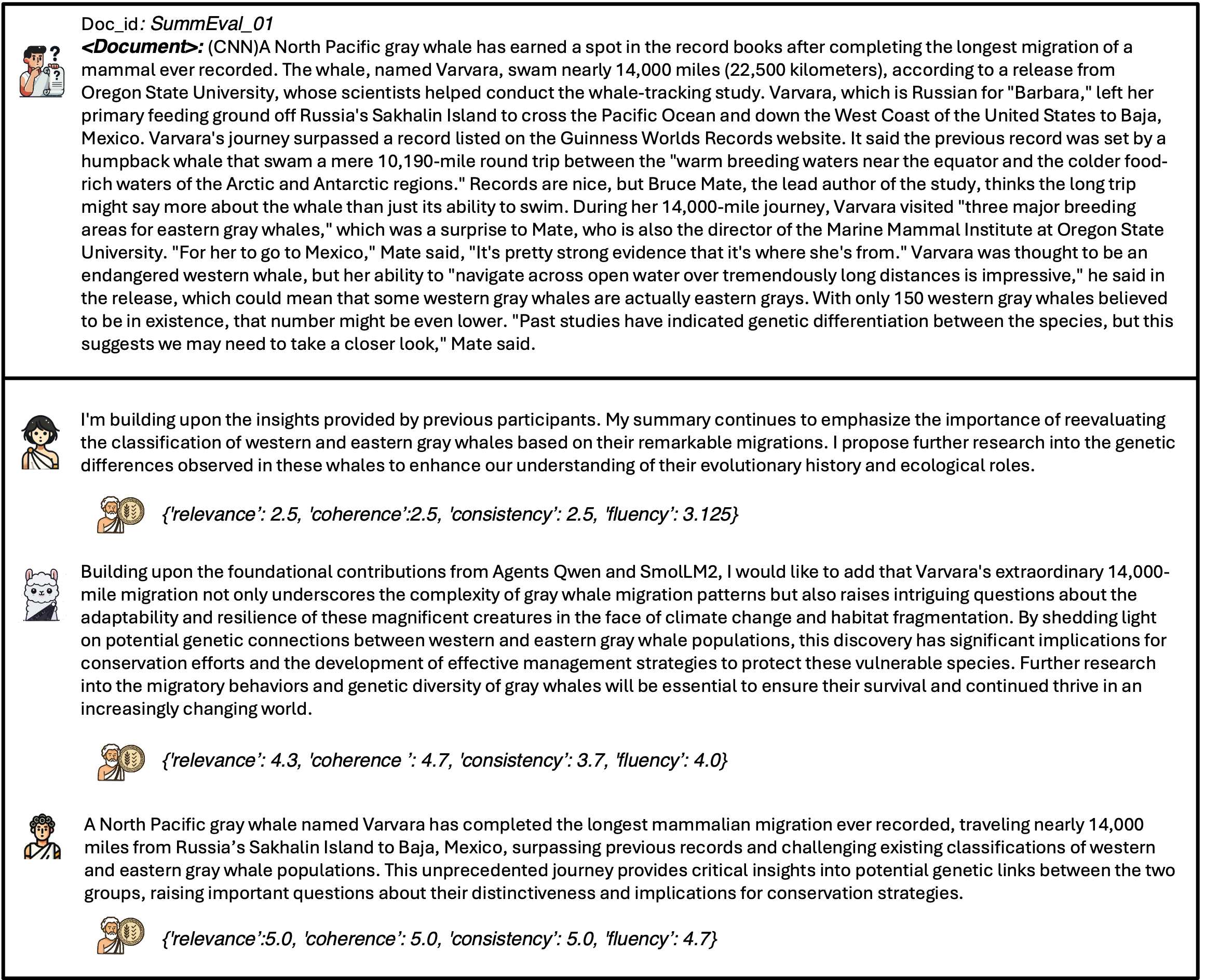}
    \caption{SummEval Case Study. A summarization task where the initial response lacks cohesion and informativeness. Subsequent agents improve sentence structure, factual completeness, and coherence. Evaluations at each stage are conducted by Qwen2.5-7B-Instruct (frozen evaluator). The final summary exhibits significantly enhanced quality as judged by the evaluator, confirming the utility of MAC-SPGG in generation tasks.}
    \label{fig:case3}
\end{figure}

To qualitatively illustrate the collaborative dynamics fostered by our MAC-SPGG framework, we present three representative case studies in Figures~\ref{fig:case2} and \ref{fig:case3}. These examples involve a diverse ensemble of large language models (LLMs), including Qwen3-8B, SmolLM2-1.7B-Instruct, LLaMA3.1-8B-Instruct, and Qwen2.5-7B-Instruct. Among these, Qwen2.5-7B-Instruct is used as a \textit{trained evaluator}, which is fine-tuned for contribution assessment tasks and kept frozen during inference (i.e., it does not generate content or update parameters). See Appendix~\ref{app-exp-rl} for training details. The remaining models function as sequential contributors, collaboratively refining the output through the MAC-SPGG protocol. To ensure computational efficiency and compatibility with limited GPU memory, all models are deployed using 8-bit quantization.

\vspace{1em}

\vspace{1em}

These case studies highlight MAC-SPGG’s capacity to integrate diverse models into a structured collaboration framework, facilitating improvement over time even when the individual models are imperfect. This collaborative mechanism proves effective across both reasoning-intensive (MMLU) and generation-intensive (SummEval) tasks, showcasing the generality and extensibility of the proposed approach.

\end{document}